\documentclass[11pt]{article}
\usepackage{booktabs}       
\usepackage{amsfonts}       
\usepackage{amsmath}
\usepackage{graphicx}
\usepackage{amsthm}
\usepackage{amssymb}
\usepackage{multirow}
\usepackage{makecell}
\usepackage[vlined,linesnumbered,ruled,resetcount]{algorithm2e}
\usepackage[round]{natbib}
\usepackage[colorlinks,linkcolor=magenta,filecolor=blue,citecolor=blue,urlcolor=blue]{hyperref}%
\usepackage[top=1in, left=1in, right=1in, bottom=1in]{geometry}
\usepackage{subfigure}
\usepackage{amssymb}
\usepackage{pifont}
\usepackage{mathtools}
\usepackage{stmaryrd}
\usepackage[T1]{fontenc}

\begin{document}

\title{\bf\Huge Package for Fast ABC-Boost}

\author{\textbf{Ping Li} and \textbf{Weijie Zhao}\\\\
Cognitive Computing Lab\\
Baidu Research\\
10900 NE 8th St. Bellevue, WA 98004, USA\\\\
\texttt{\{pingli98, zhaoweijie12\}@gmail.com}
}

\date{June 2022}

\maketitle              
\begin{abstract}
\noindent This report presents the open-source package~\url{https://github.com/pltrees/abcboost} which implements the series of boosting works over the past many years~\citep{li2008adaptive,li2009abc,li2010fast,li2010robust,li2022fast,li2022pGMM}. In particular, this package includes mainly \textbf{three} lines of techniques, among which the following two techniques are already the standard implementations in popular boosted tree platforms: 
\begin{itemize}
    \item[(i)] The histogram-based (feature-binning) approach makes the tree implementation convenient and efficient. In~\citet{li2007mcrank}, a simple fixed-length adaptive binning algorithm was developed. In this report, we demonstrate that such a simple algorithm is still surprisingly effective compared to  more  sophisticated variants in popular tree platforms. 
\item[(ii)] The explicit gain formula~\citep{li2010robust} for tree splitting based on second-order derivatives of the loss function typically improves, often considerably, over the first-order methods. Although the gain formula in~\citet{li2010robust} was derived for logistic regression loss, it is indeed a generic formula for loss functions with second-derivatives. For example, the open-source package also includes $L_p$ regression for general $p\geq 1$, not limited to just $p=2$ or $p=1$. 
\end{itemize}

\vspace{0.1in}

\noindent The main contribution of this package is the ABC-Boost (adaptive base class boosting) for multi-class classification. The  initial work~\citep{li2008adaptive} derived a new set of derivatives of the classical multi-class logistic regression loss function by specifying a ``base class''. The classification accuracy can be substantially improved if the base class is chosen properly. The major technical challenge is to design a search strategy to select the  base class efficiently and effectively. The prior published works in~\citet{li2009abc,li2010robust} implemented an exhaustive search procedure to find the base class at each iteration which is computationally too expensive. More efficient search strategies were developed  in~\citet{li2008adaptive,li2010fast} which were not formally published. Recently, a new  report~\citep{li2022fast} presents a unified framework of ``Fast ABC-Boost'' by introducing a ``search'' parameter, a ``gap'' parameter, and a ``warm-up'' parameter. These parameters allow users to flexibly and efficiently choose the proper search space for the base class. \citet{li2022fast} has demonstrated the excellent empirical performance of this unified framework.

\vspace{0.1in}

\noindent We hope this open-source packages would benefit machine learning practitioners. In our experience, boosted trees are still highly effective in numerous practical scenarios, except perhaps for applications (e.g., ads CTR models) with extremely high-dimensional sparse features (e.g., hundreds or thousands of billions of features with merely hundreds of non-zero features).

\vspace{0.1in}

\noindent \textbf{The package provides interfaces for linux, windows, mac, matlab, R, python,~etc.} 

\noindent 

\end{abstract}

\newpage\clearpage

\section{Introduction}

\noindent The ``Fast ABC-Boost'' package is the effort of more than a dozen years of work, at Cornell University, Rutgers University, and Baidu Research. See for example the lecture notes at Cornell and Rutgers: 
\begin{itemize}
\item {\footnotesize\url{http://statistics.rutgers.edu/home/pingli/STSCI6520/Lecture/ABC-LogitBoost.pdf}} .
\item {\small\url{http://www.stat.rutgers.edu/home/pingli/doc/PingLiTutorial.pdf}} (pages 15--77).
\end{itemize}

Some machine learning researchers might still recall the  discussions  \url{https://hunch.net/?p=1467}  in 2010, on the paper by~\citet{li2010robust} which developed ``Robust LogitBoost'' and ``Adaptive Base Class Boost''. At that time, researchers were curious why boosted trees were so effective compared to deep neural networks on the  datasets developed by deep learning researchers~\citep{larochelle2007empirical}. 

A decade has passed since that discussion. Needless to say, deep neural networks have achieved overwhelming success in numerous fields of research and practice. After developing boosted tree algorithms in~\citet{li2007mcrank,li2008adaptive,li2009abc,li2010fast,li2010robust}, the author(s) moved to many other interesting research topics including randomized sketching/hashing methods (e.g.,~\citet{li2022gcwsnet}), approximate near neighbor search \& neural ranking (e.g.,~\citet{tan2021fast}), deep neural networks \& approximate near neighbor search for advertising (e.g.,~\citet{fan2019mobius,fei2021gemnn}), GPU architecture for massive-scale CTR models (e.g.,~\citet{zhao2022communication}, also see the media report {\small\url{www.nextplatform.com/2021/06/25/a-look-at-baidus-industrial-scale-gpu-training-architecture}}),  ads CTR model compression (e.g.,~\citet{xu2021agile}), AI model security (e.g.,~\citet{zhao2022integrity}), privacy, theory, etc., as well as machine learning applications in NLP, knowledge graphs, and vision. 

\vspace{0.1in}

Nonetheless, despite the widespread success of deep neural networks, we have found that boosted trees are still extremely useful in practice such as ranking of search results, stock price prediction, finance risk models, and much more. For example, we used boosted trees for Baidu's input method editor (IME) and deployed tree models on cell phones~\citep{wang2020improved}. In our own experience, we find boosted trees are well-suited for prediction tasks which have (e.g.,) less than 10000 features and (e.g.,) less than 100 million training examples. For applications (such as ads CTR predictions) using extremely high-dimensional sparse data with hundreds of billions of training examples, typically deep neural networks are more convenient or more effective. 

\vspace{0.1in}

As always, we should first solute to pioneers in boosting and trees,  e.g.,~\citet{brieman1983classification,schapire1990strength,freund1995boosting,freund1997decision,bartlett1998boosting,schapire1999improved,friedman2000additive,friedman2001greedy}. As summarized in a recent paper on merging decision trees~\citep{fan2020classification}, in the past 15 years or so, multiple practical developments have enhanced the performance as well as the efficiency of boosted tree algorithms, including
\begin{itemize}
    \item The explicit (and robust) formula for tree-split criterion using the second-order gain information~\citep{li2010robust} (i.e., ``Robust LogitBoost'') typically improves the accuracy, compared to the implementation based on the criterion of using only the first-order gain information~\citep{friedman2001greedy}. It is nowadays the standard implementation in popular tree platforms. 

\item The adaptive binning strategy developed in~\cite{li2007mcrank} effectively transformed  features to integer values and substantially simplified the implementation and improved the efficiency of trees as well. Binning is also the standard implementation of popular tree platforms. 

\item The ``adaptive base class boost'' (ABC-Boost) scheme~\citep{li2008adaptive,li2009abc,li2010fast,li2010robust,li2022fast} for multi-class classification, by re-writing the derivatives of the classical multi-class logistic regression loss function, often improves the accuracy of multi-class classification tasks, in many cases substantially so.
\end{itemize}

The open-source package at~\url{https://github.com/pltrees/abcboost}  includes the documentation to assist users to install the package and use it for regression,  classification, and ranking. We have compared the results on regression and  classification with two popular boosted tree platforms, i.e., LightGBM and xgboost, and notice some discrepancy in accuracy. This observation is interesting (and  might be confusing too) because they essentially implemented the same algorithm: (i) the feature binning (histogram building) before training as in~\citet{li2007mcrank}; and (ii) second-order gain information formula for tree splitting as derived in~\citet{li2010robust}. How can the implementation of the same algorithm output noticeably different results? 

\vspace{0.1in}

We realize that the discrepancy might be caused by the difference in implementing the feature binning procedure. \cite{li2007mcrank} designed an overly simplistic \textbf{fixed-length} binning method, while LightGBM and xgboost appear to use much more refined procedures. Perhaps counter-intuitively, our experiments show that the very simple binning method in~\cite{li2007mcrank} produces more accurate results on the datasets we have tested. 


Thus, in this report, we first describe the simple binning method used in the ABC-Boost package, then we demonstrate how to use ABC-Boost for regression, binary classification, and multi-class classification. For each task, we also report the experimental results of LightGBM and xgboost, based on their newest versions in June 2022.

\section{Fixed-Length Binning Method for Feature Preprocessing} 

Regression tree~\citep{brieman1983classification} is the basic building block, for classification, regression, and ranking tasks. The idea of trees is to recursively divide the data points on an axis-aligned fashion based on some ``gain'' criterion, and report the average (or weighed average) responses of the data points in the final (sub-divided) regions as the prediction values. The sub-divided regions can be organized/viewed as a tree, with the leaf nodes corresponding to the final sub-divided regions.

\begin{figure}[b!]
\vspace{-0.15in}
    \centering
    \includegraphics[width=2.7in]{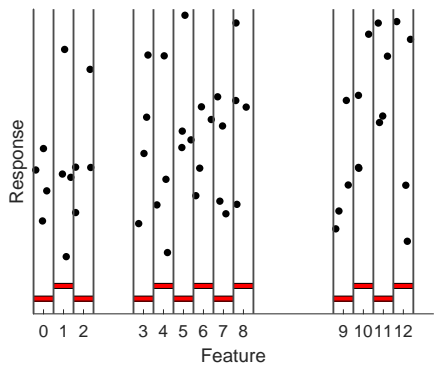}
    \vspace{-0.15in}
    \caption{Illustration of the fixed-length binning method used in this package. For this feature, the values are grouped into 13 bins, i.e., the original feature values become integers between 0 to 12.  }
    \label{fig:tree-bin}\vspace{-0.1in}
\end{figure}

Therefore, a crucial task is to compute the best split point for each feature and choose the best feature  (i.e., the feature with the largest gain) to conduct the actual split by dividing the data points in the current node into two parts. This procedure continues recursively until some stopping criterion is met. Before~\citet{li2007mcrank}, typical tree implementations first sort the data points for each dimension according to the feature values and need to keep tracks of the data points after splitting. As shown in Figure~\ref{fig:tree-bin}, \citet{li2007mcrank} first quantize (bin) the feature values to be integers, which are naturally  ordered. This trick has simplified the implementation very considerably. It can also make the procedure more efficient if the total number of bins is not too large. The binning in Figure~\ref{fig:tree-bin} also adapts to the data distribution as it only assigns bin values where there are data.

\newpage\clearpage

As shown in the following matlab code, the binning method is extremely simple: We first start with a very small initial bin-length (e.g., $10^{-10}$)  and pre-specified ``MaxBin'' parameter such as 128 or 1024. For each feature, we first sort the data points according to the feature values. We assign bin numbers to data points from the smallest to the largest, wherever there are data points, until the number of bins needed exceeds MaxBin. Then we start over  by doubling the bin-length.

This binning procedure has many obvious advantages. It is very simple and easy to implement. It naturally adapts to data distributions. This procedure does not impact features which are already ordered categorical variables. The disadvantages of this procedure are also obvious. It is by no means an ``optimal'' algorithm in any sense and we expect it could be improved in many ways. For example, using fixed-length, we would expect to see poor performance if MaxBin is set to be too small such as 10. On the other hand, the tree algorithm itself would not work well any way if MaxBin is set to be too small. As shown in the experiments later in this report, this extremely simple binning method performs very well for trees. We provide the matlab code to help readers to better understand the procedure and help researchers to potentially improve their binning algorithms (and tree platforms).

{\footnotesize
\begin{verbatim}
function adabin(inputfile, max_bin)

%%%%%%%%%%%%%%%%%%%%%%%%%%%%%%%%%%%%%%
%%%%  adaptive binning algorithm  %%%%
%%%%%%%%%%%%%%%%%%%%%%%%%%%%%%%%%%%%%%
% Quantizing input data to at most "max_bin" values. In this example, we quantize the input matrix 
% starting from the second column because the first column is assumed to be the labels.  

input = feval('load',inputfile);
output = input; 
for i = 2:size(input,2)    
    col = adabin1feature(input(:,i),max_bin);
    output(:,i) = col; 
end
outputfile = [inputfile '.bin' num2str(max_bin) '.csv'];
writematrix(output,outputfile);


function output = adabin1feature(col,max_bin)

bin_len = 1e-10;
[data, ind] = sort(col);
output = data;
while(1)
    cur_bin = 0; cur_ind = 1;
    for i = 1:size(data,1)
        if(data(i)-data(cur_ind)>bin_len)
            cur_bin = cur_bin+1;
            cur_ind = i;
            if(cur_bin>max_bin)
                bin_len = bin_len*2;
                break;
            end
        end
        output(i) = cur_bin;
    end
    if(cur_bin<=max_bin)
        break;
    end
end
output(ind) = output;    
\end{verbatim}
}

In summary, this seemingly too simple (fixed-length) binning algorithm works well for boosted tree methods, likely due to two main reasons:
\begin{itemize}

\item The maximum allowed number of bins (i.e., the MaxBin parameter) should not be too small any way for boosted trees. Too much information would be lost if the data are too coarsely quantized. With that many bins (e.g., MaxBin = 1000), it is probably not so easy to improve this fixed-length strategy, as far as the performance of booting trees is concerned. 

\item We  should not expect all features  would use the same number of bins. Typically, in one dataset, the features can differ a lot. For example, some features might be binary  (i.e., even  using MaxBin = 1000 would only generate two values), some features may have just 100 distinct  values (i.e., using MaxBin = 1000 would still just generate at most 100 values), and some features really need more quantization levels. Therefore, the parameter MaxBin is just a crude guideline. Trying too hard to ``optimize'' the binning procedure according to a given MaxBin is likely counter-productive. 
\end{itemize}

In practice, we recommend setting MaxBin = 100 (or 128) as a starting point. If the performance is not satisfactory, we can gradually increase it to (e.g.,) MaxBin = 1000 (or 1024). In our experience, it is quite rare to observe noticeably much better performance once MaxBin  is larger than 1000. 

\vspace{0.1in}

In the next three sections, we will present experimental results on regression, binary classification, and multi-class classification, using the Fast ABC-Boost package. We will compare the results with LightGBM and xgboost by varying MaxBin from $10$ to $10^4$, to illustrate the impact of MaxBin on the performance. Also, we notice that the results become non-deterministic once we turn on multi-threading, although the stochastic variations are usually not too large. In order to strictly ensure deterministic results (for clear comparisons), we run all the experiments as single-thread. 

\section{$L_p$ Regression}

Readers please refer to a  detailed report on $L_p$ boosting for regression~\citep{li2022pGMM}.  Consider a training dataset  $\{y_i,\mathbf{x}_i\}_{i=1}^n$, where $n$ is the number of samples, $\mathbf{x}_i$ is the $i$-th feature, and  $y_i$ is the $i$-th value. The goal of $L_p$ regression  is to build a model $F(\mathbf{x})$ to minimize the $L_p$ loss:
\begin{align}\label{eqn:Lp}
L_p = \frac{1}{n}\sum_{i=1}^n L_{i} = \frac{1}{n}\sum_{i=1}^n |y_i - F_i|^p, \hspace{0.2in}\text{where } F_i = F(\mathbf{x}_i).
\end{align}
Using the ``additive model''~\citep{friedman2000additive,friedman2001greedy},  we let $F$ be a sum of $M$ terms:
\begin{align}\label{eqn_F_M}
F(\mathbf{x}) = F^{(M)}(\mathbf{x}) = \sum_{m=1}^M f_m(\mathbf{x}),
\end{align}
where  $f_m(\mathbf{x})$, the base learner, is  a regression tree and learned from the data in a stagewise greedy fashion. Following the idea from~\citet{friedman2000additive}, at each boosting iteration, we fit $f_m$ by weighted least squares, with  responses $\{z_i\}$ and weights $\{w_i\}$: 
\begin{align}\label{eqn:zw}
&z_i = \frac{-L_{i}^\prime}{L_{i}^{\prime\prime}},\hspace{0.2in} w_i = L_{i}^{\prime\prime},\hspace{0.2in} \\
\text{where} \ \ & L_{i}^\prime = \frac{\partial{L_{i}}}{\partial F_i} =-p|y_i-F_i|^{p-1}\text{sign}\left(y_i-F_i\right),\hspace{0.2in} (p\geq 1)\\
&L_{i}^{\prime\prime} = \frac{\partial^2{L_{i}}}{\partial F_i^2} =  p(p-1)|y_i-F_i|^{p-2},\hspace{0.2in} (p\geq 2)
\end{align}

\cite{li2010robust} derived the corresponding \textbf{gain} formula  needed for deciding the split location in building regression trees using  responses $\{z_i\}$ and weights $\{w_i\}$.  Historically, boosting based on the weighted least square procedure  was believed to suffer from numerical issues~\citep{friedman2000additive,friedman2008response},  and hence later~\citet{friedman2001greedy} proposed using only the first derivatives to fit the trees, i.e., 
\begin{align}\label{eqn:zw1}
z_i = -L_{p,i}^\prime, \hspace{0.2in} w_i=1. 
\end{align}
It is now clear that, as shown in~\cite{li2010robust}, one can derive the explicit and numerically stable/robust formula for computing the gains using second-order information.

\subsection{Tree-Splitting Criterion Using Second-Order Information}\label{sec_split}

Consider a tree node with $N$ data points and one particular feature.  We have the weights $w_i$ and the response values $z_i$, $i=1$ to $N$. The data points are already sorted according to the sorted order of the feature values. The tree-splitting procedure is to find the index $s$, $1\leq s<N$, such that the weighted  square error (SE) is reduced the most if split at $s$.  That is, we seek the $s$ to maximize
\begin{align}\notag
Gain(s) = &SE_{total} - (SE_{left} + SE_{right})\\\notag
=&\sum_{i=1}^N (z_i - \bar{z})^2w_i - \left[
\sum_{i=1}^s (z_i - \bar{z}_L)^2w_i + \sum_{i=s+1}^N (z_i - \bar{z}_R)^2w_i\right],
\end{align}
where $\bar{z} = \frac{\sum_{i=1}^N z_iw_i}{\sum_{i=1}^N w_i}$,
$\bar{z}_{left} = \frac{\sum_{i=1}^s z_iw_i}{\sum_{i=1}^s w_i}$,
$\bar{z}_{right} = \frac{\sum_{i=s+1}^N z_iw_i}{\sum_{i=s+1}^{N} w_i}$.  With some algebra, we can obtain
\begin{align}\notag
Gain(s) =& \frac{\left[\sum_{i=1}^s z_iw_i\right]^2}{\sum_{i=1}^s w_i}+\frac{\left[\sum_{i=s+1}^N z_iw_i\right]^2}{\sum_{i=s+1}^{N} w_i}- \frac{\left[\sum_{i=1}^N z_iw_i\right]^2}{\sum_{i=1}^N w_i}
\end{align}
Plugging in  $z_i = -L_i^{\prime}/L_i^{\prime\prime}$, $w_i = L_i^{\prime\prime}$  yields,
\begin{align}\label{eqn:gain}
Gain(s) =&  \frac{\left[\sum_{i=1}^s L_i^\prime \right]^2}{\sum_{i=1}^s L_i^{\prime\prime}}+\frac{\left[\sum_{i=s+1}^N L_i^{\prime}\right]^2}{\sum_{i=s+1}^{N} L_i^{\prime\prime}}- \frac{\left[\sum_{i=1}^N  L_i^\prime \right]^2}{\sum_{i=1}^N L_i^{\prime\prime}}.
\end{align}
This procedure is numerically robust/stable because we never need to directly compute the response values $z_i =- L_i^\prime/L_i^{\prime\prime}$, which can (and should) approach infinity easily. Because the original LogitBoost~\citep{friedman2000additive} used the individual response values  $z_i =- L_i^\prime/L_i^{\prime\prime}$, the procedure was believed to have numerically issues, which was one motivation for~\cite{friedman2001greedy} to use only the first derivatives to build tress i.e., $z_i = L_i^\prime$ $w_i =1$.  Thus the gain formula becomes
\begin{align}\label{eqn:ugain}
UGain(s) =&  \frac{1}{s}\left[\sum_{i=1}^s L_i^\prime \right]^2+
\frac{1}{N-s}\left[\sum_{i=s+1}^N  L_i^\prime \right]^2-\frac{1}{N}
\left[\sum_{i=1}^N  L_i^\prime \right]^2.
\end{align}

\subsection{$L_p$ Boosting Algorithm }

Algorithm~\ref{alg:LpBoost} describes $L_p$ boosting for regression  using the tree split gain formula~\eqref{eqn:gain} (for $p\geq 2$) or the tree split gain formula~\eqref{eqn:ugain} (for $1\leq p<2$). Note that after trees are constructed, the values of the terminal nodes are computed by
\begin{align}\notag
\frac{\sum_{node} z_{i,k} w_{i,k}}{\sum_{node} w_{i,k}} =
\frac{\sum_{node}-L_i^\prime}{\sum_{node} L_i^{\prime\prime}},
\end{align}
which explains Line 5 of Algorithm~\ref{alg:LpBoost}.  When $1\leq p<2$,  we follow~\citet{friedman2001greedy} by  using the first derivatives to build the trees with the split gain formula~\eqref{eqn:ugain}, and  update the terminal node~as
\begin{align}\notag
\frac{\sum_{node} z_{i,k} w_{i,k}}{\sum_{node} w_{i,k}} =
\frac{\sum_{node}-L_i^\prime}{p \times \#|node|}.
\end{align}

{\begin{algorithm}[t]
$F_{i} = 0$, $i = 1$ to $n$ \\
For $m=1$ to $M$ Do\\
\hspace{0.15in}  If $p\geq 2$ Do\\
\hspace{0.3in}  $\left\{R_{j,m}\right\}_{j=1}^J = J$-terminal node weighted regression tree from
 $\{z_i = -L_i^\prime/L_i^{\prime\prime}, \ \ w_i = L_i^{\prime\prime}, \ \  \mathbf{x}_{i}\}_{i=1}^n$, using the tree split gain formula Eq.~\eqref{eqn:gain}.\\
 \hspace{0.3in}  $\beta_{j,m} = \frac{ \sum_{\mathbf{x}_i \in
  R_{j,m}} -L_i^\prime}{ \sum_{\mathbf{x}_i\in
  R_{j,m}} L_i^{\prime\prime} }$ \\
 \hspace{0.15in} End\vspace{0.1in}\\
\hspace{0.15in}  If $1\leq p< 2$ Do \\
\hspace{0.3in}  $\left\{R_{j,m}\right\}_{j=1}^J = J$-terminal node regression tree from
 $\{z_i = -L_i^\prime, \ \  \mathbf{x}_{i}\}_{i=1}^n$, using the tree split gain formula Eq.~\eqref{eqn:ugain}.\\
 \hspace{0.3in}  $\beta_{j,m} = \frac{ \sum_{\mathbf{x}_i \in
  R_{j,m}} -L_i^\prime}{ 
  p\times \#|R_{j,m}|}$ \\  
 \hspace{0.15in} End\\  
\hspace{0.15in}  $F_{i} = F_{i} +
\nu\sum_{j=1}^J\beta_{j,m}1_{\mathbf{x}_i\in R_{j,m}}$ \\
End
\caption{$L_p$ boosting. $L^\prime = -p|y_i-F_i|^{p-1}\text{sign}\left(y_i-F_i\right)$,  $L^{\prime\prime} = p(p-1)|y_i-F_i|^{p-2}$.}
\label{alg:LpBoost}
\end{algorithm}}

\vspace{-0.2in}
\subsection{Experiments}

Following the instruction on \url{https://github.com/pltrees/abcboost} , users can install Fast ABC-Boost package. Assume the executables are on the current directory and datasets are available on the ``data/'' directory.  The ``comp-cpu'' dataset, available in both libsvm and csv formats, has 4096 examples for training and 4096 examples for testing.  From the  terminal, the following command 

\vspace{-0.1in}
{\scriptsize
\begin{verbatim}
./abcboost_train -method regression -lp 2 -data data/comp_cpu.train.csv -J 20 -v 0.1 -iter 10000 -data_max_n_bins 1000
\end{verbatim}
}
\noindent builds an $L_2$ regression boost model with $J=20$ leaf nodes and $\nu = 0.1$ shrinkage, for a maximum of 10000 iterations. The maximum number of bins (MaxBin) is set to be 1000. We adopts a conservative early stopping criterion and let program exit after the $L_p$ loss is lower than
\begin{align}\label{eqn:stop}
\epsilon^{p/2} \times \frac{1}{n}\sum_{i=1}^n{|y_i|^p},
\end{align}
where $\epsilon = 10^{-5}$ by default. In this example, the program exits after 933 iterations (instead of 10000 iterations). After training, two files are created on the current directory: 
\begin{verbatim}
comp_cpu.train.csv_regression_J20_v0.1_p2.model  
comp_cpu.train.csv_regression_J20_v0.1_p2.trainlog
\end{verbatim}

\vspace{0.1in}

To test the trained model on the test dataset, we run 
{\scriptsize
\begin{verbatim}
./abcboost_predict -data data/comp_cpu.test.csv -model comp_cpu.train.csv_regression_J20_v0.1_p2.model 
\end{verbatim}
}
\noindent which generates two more (text) files to store the testing results: 
\begin{verbatim}
comp_cpu.test.csv_regression_J20_v0.1_p2.testlog
comp_cpu.test.csv_regression_J20_v0.1_p2.prediction
\end{verbatim}
\noindent The ``.testlog'' file records the test losses and other information. The ``.prediction'' stores the regression prediction values for all the testing examples at the final (or specified) iteration.

\vspace{0.1in}

\begin{figure}[h]

\vspace{-0.15in}

\begin{center}
\mbox{    
    \includegraphics[width=2.9in]{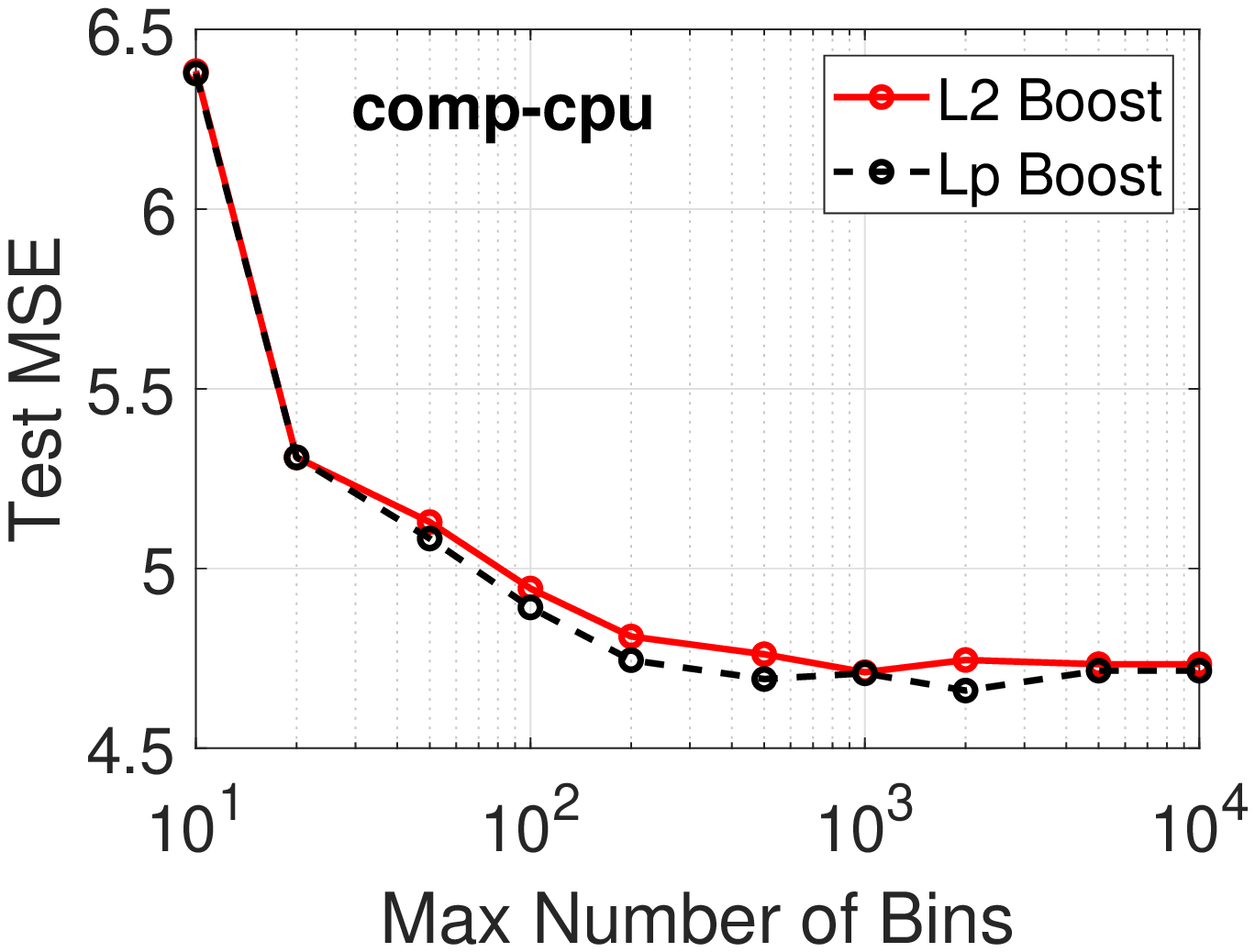}
    \includegraphics[width=2.9in]{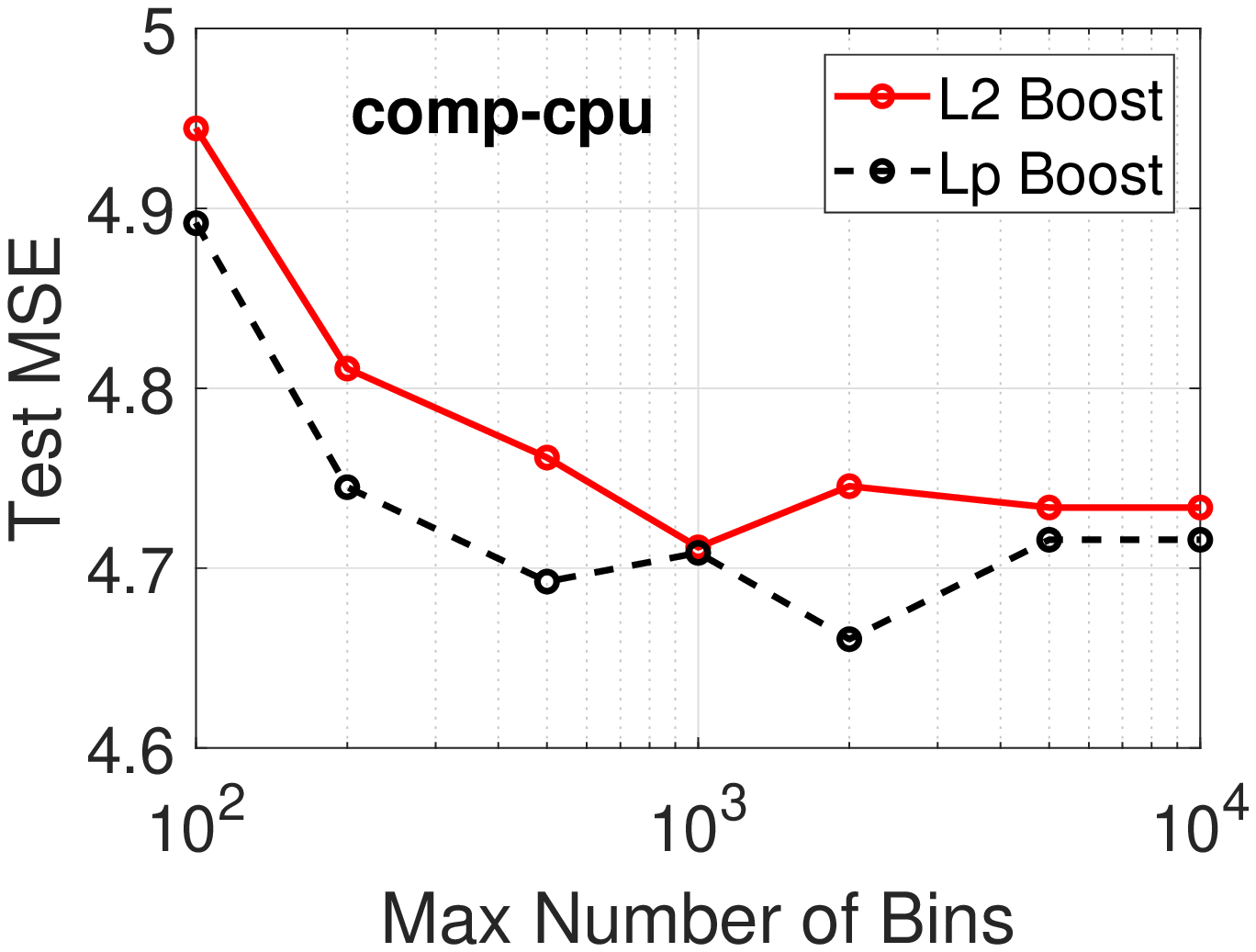}
}

\end{center}

\vspace{-0.2in}

    \caption{Best Test MSEs for $L_2$ regression (solid curve) and $L_p$ regression (dashed curve). The right panel is merely the zoomed-in version of the left panel.}
    \label{fig:MSELpL2}
\end{figure}

We experiment with parameters $J\in\{6, 10, 20\}$, $\nu\in\{0.06, 0.1, 0.2\}$,  $p$ ranging from 1 to 10, and MaxBin ranging from 10 to $10^4$. Figure~\ref{fig:MSELpL2} plots the best (among all the parameters and iterations) test MSEs at each MaxBin value. On each panel, the solid curve plots the best test MSE for $L_2$ regression and the dashed curve  for $L_p$ regression (at the best $p$). The right panel is  the zoomed-in version of the left panel to focus on MaxBin ranging from $100$ to $10^4$. For this dataset, using MaxBin = 1000 achieves good results and  using larger MaxBin values does not lead to much better results.

\begin{figure}[h]
\vspace{-0.15in}
\begin{center}
\mbox{    
    \includegraphics[width=2.9in]{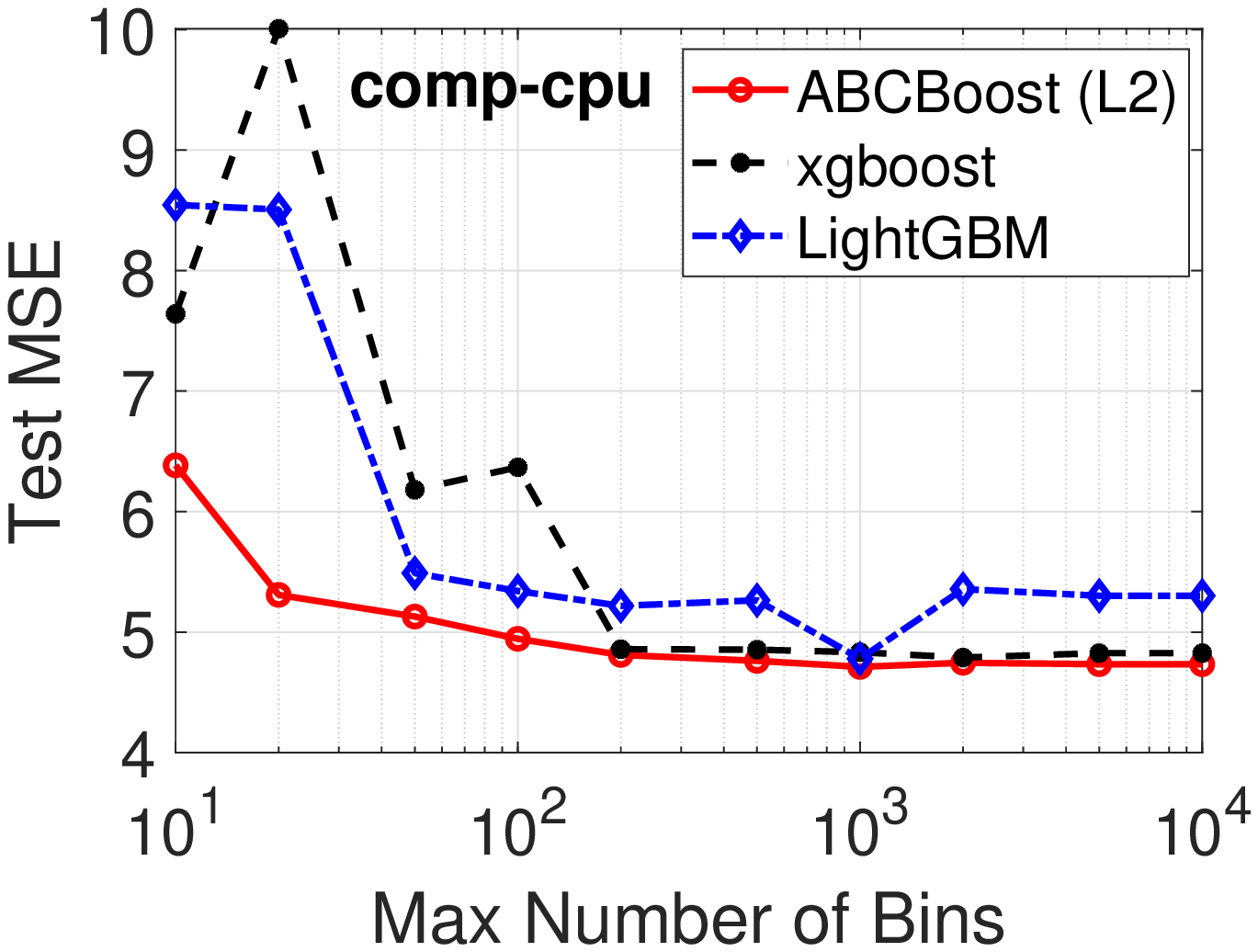}
    \includegraphics[width=2.9in]{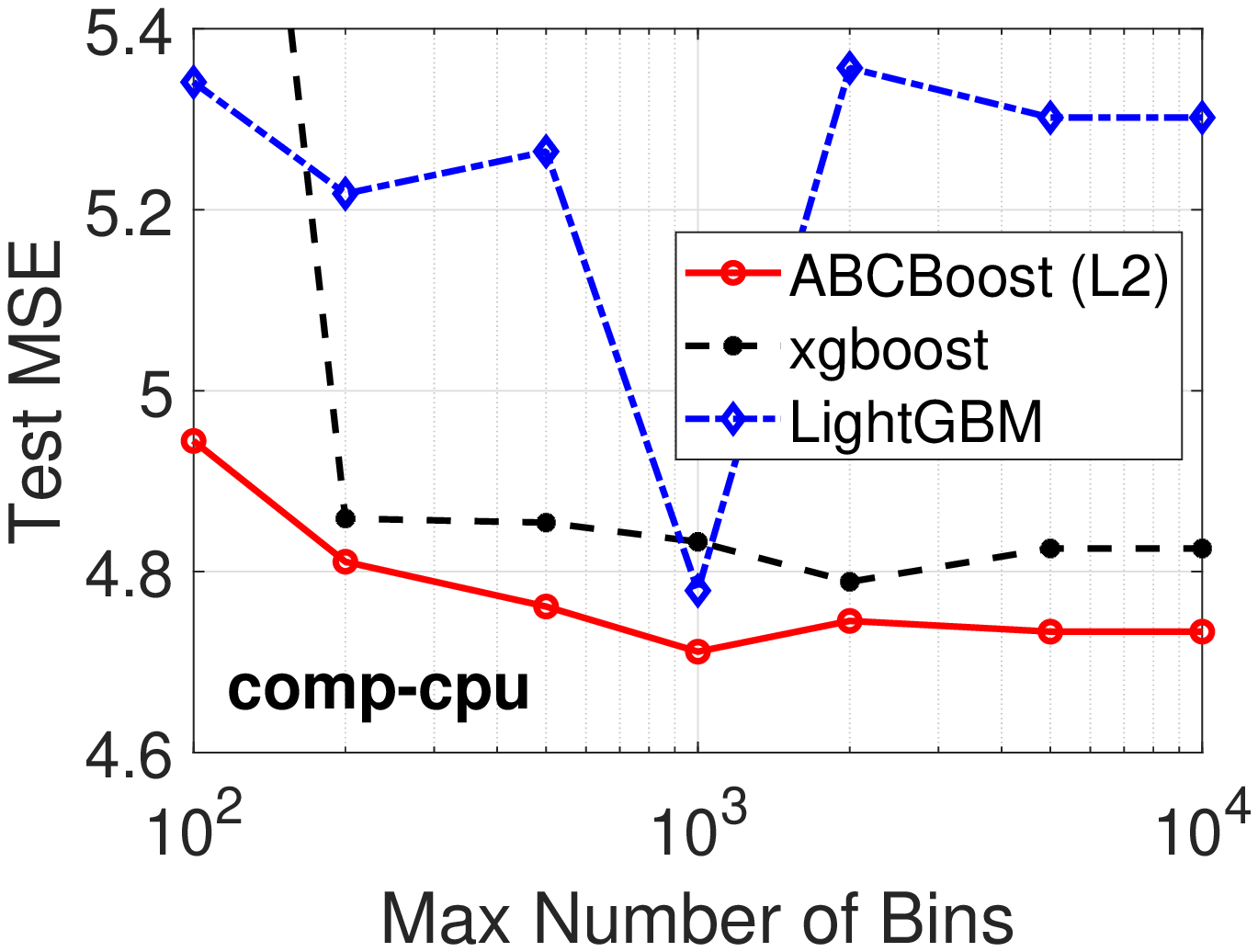}
}

\end{center}

\vspace{-0.2in}

    \caption{Best Test MSEs for $L_2$ regression based on results from three packages. }
    \label{fig:MSEThree}
\end{figure}

Figure~\ref{fig:MSEThree} plots the best test MSEs for $L_2$ regression for all three packages: ABC-Boost ($L_2$), xgboost, and LightGBM, for MaxBin ranging from 10 to $10^4$. Again, the right panel is merely the zoomed-in version of the left panel. Of course, as already shown in Figure~\ref{fig:MSELpL2}, ABC-Boost would be able to achieve even lower MSEs by using $L_p$ regression with $p\neq 2$. 

\begin{figure}[h]

\begin{center}
\mbox{    
    \includegraphics[width=2.7in]{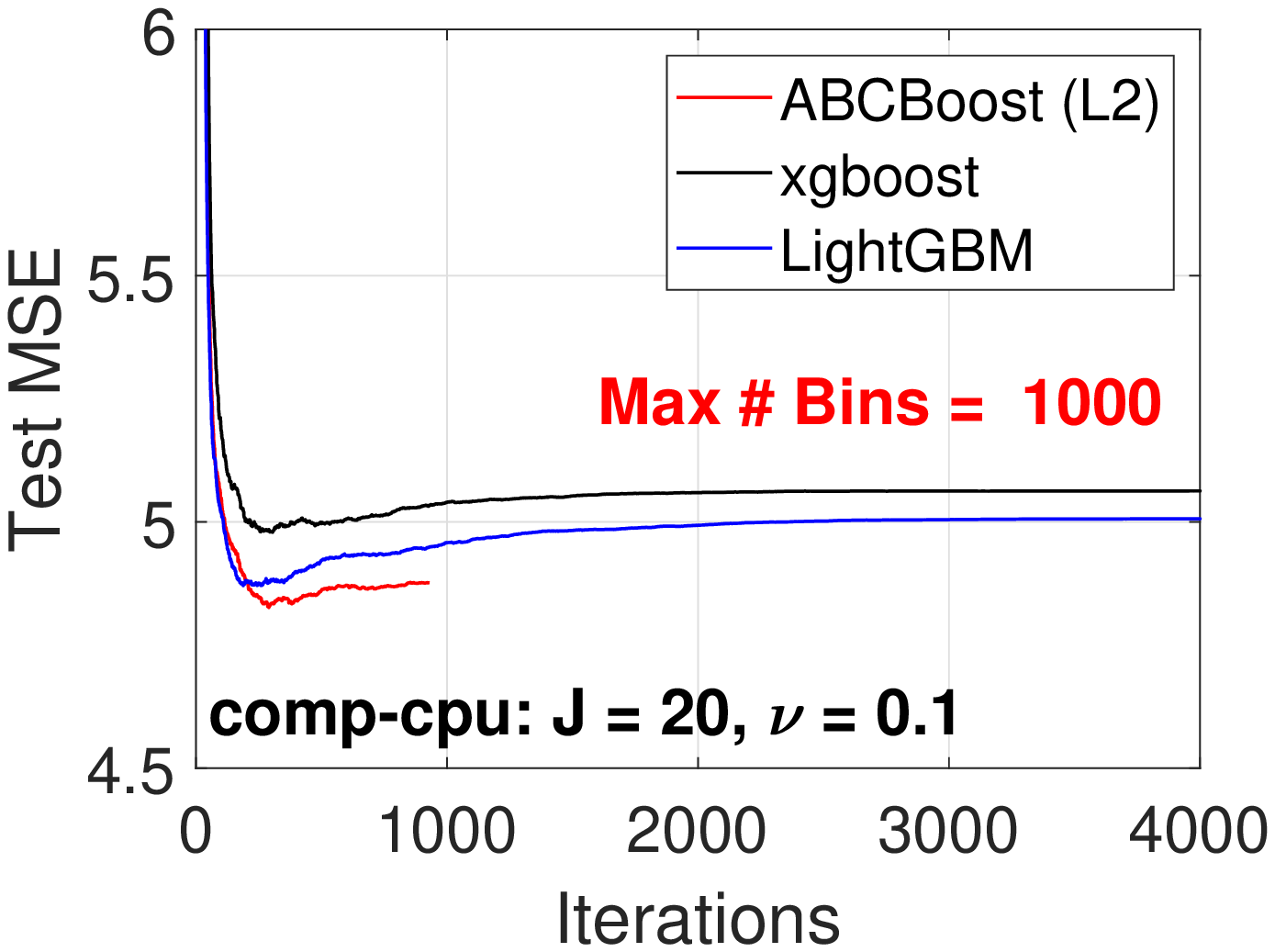}
    \includegraphics[width=2.7in]{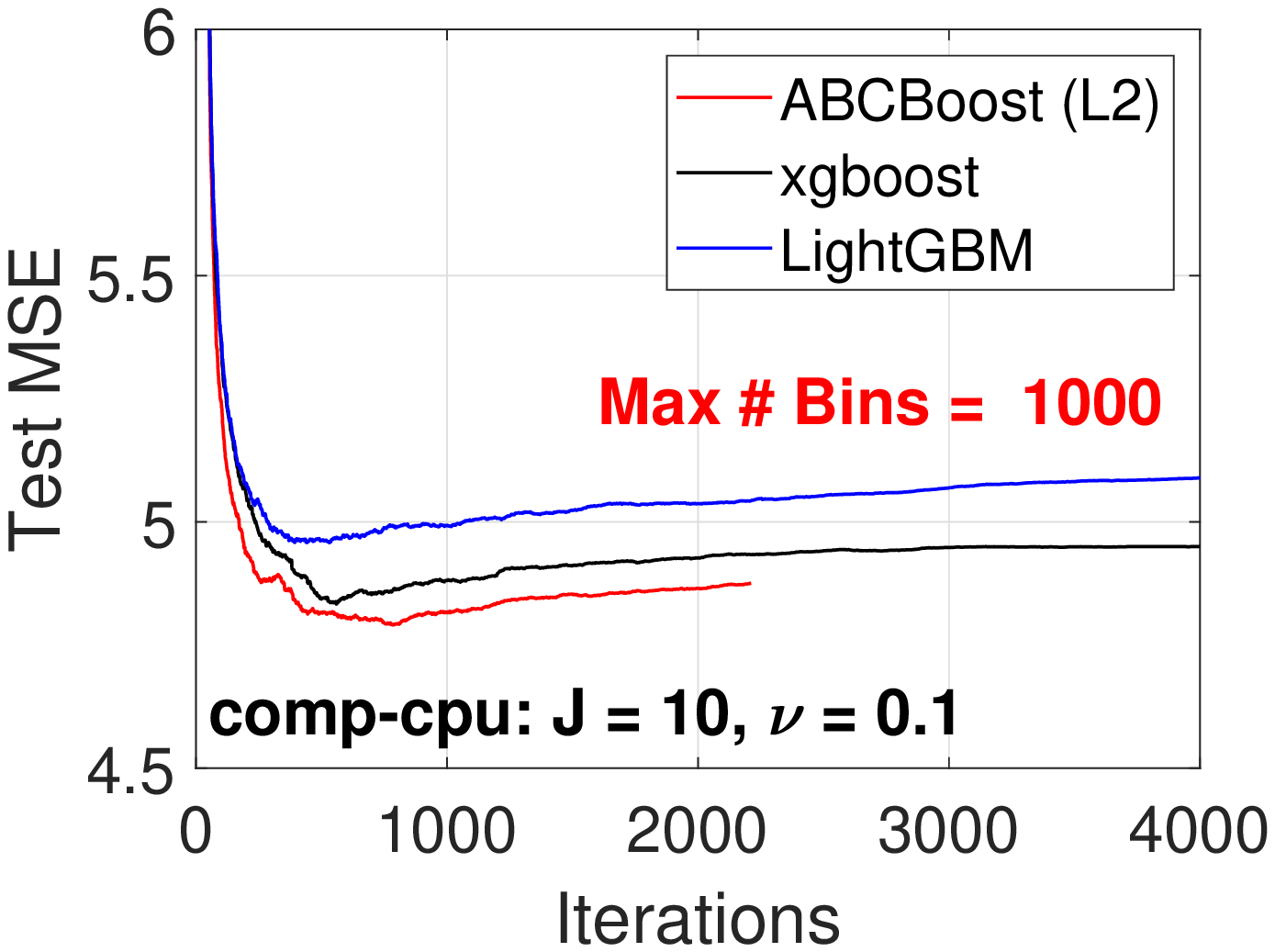}
}

\mbox{    
    \includegraphics[width=2.7in]{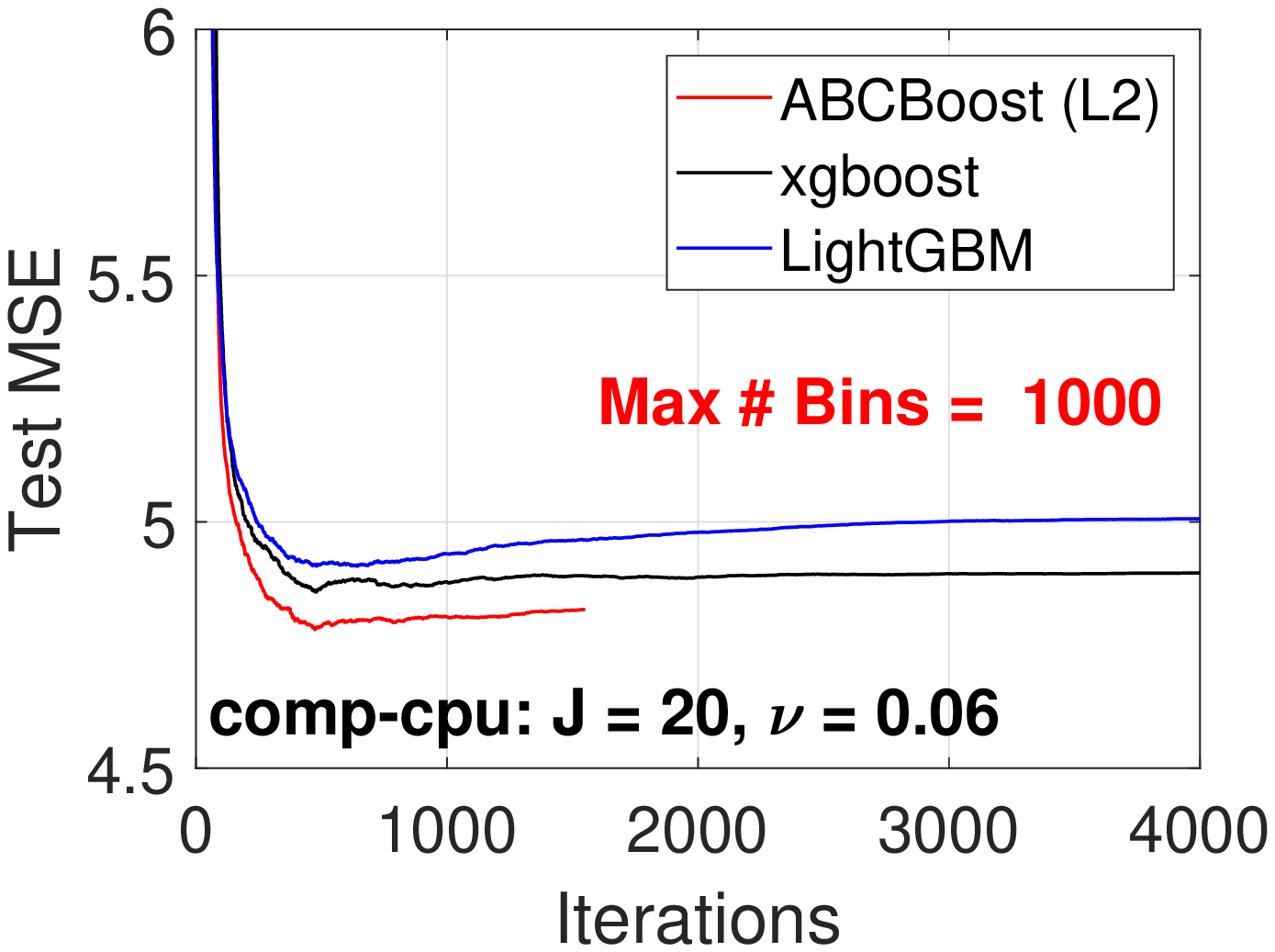}
    \includegraphics[width=2.7in]{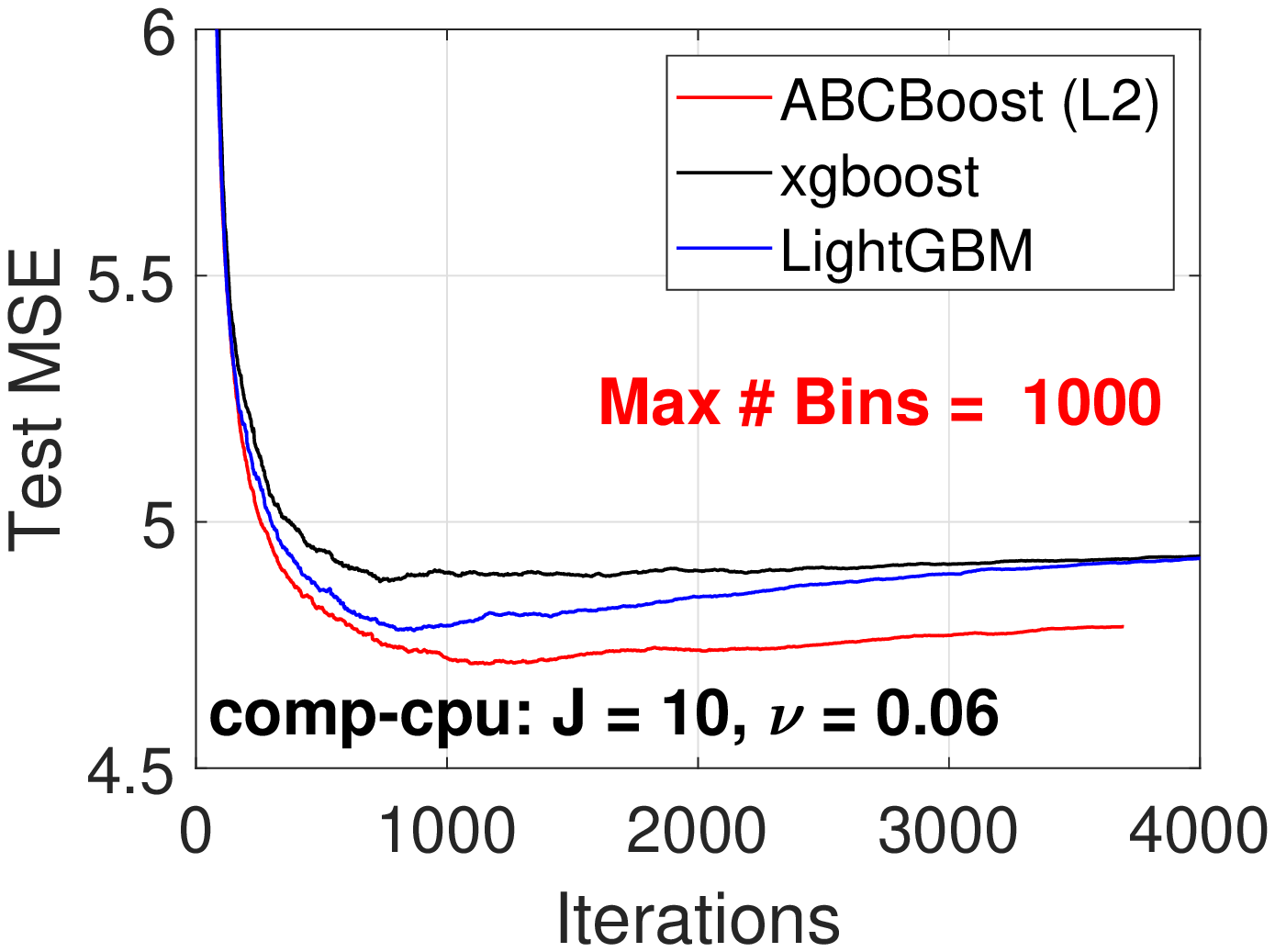}
}

\mbox{    
    \includegraphics[width=2.7in]{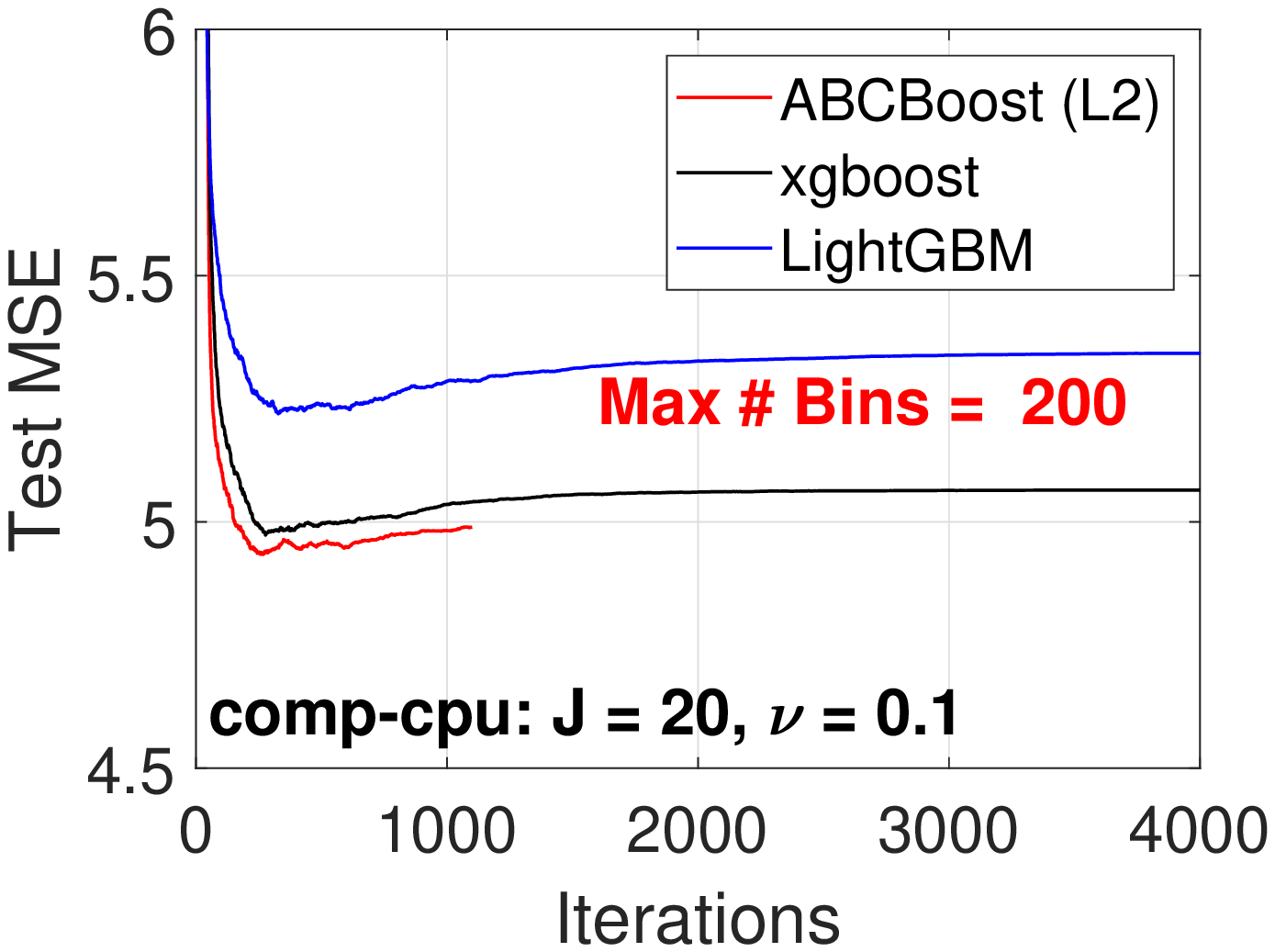}
    \includegraphics[width=2.7in]{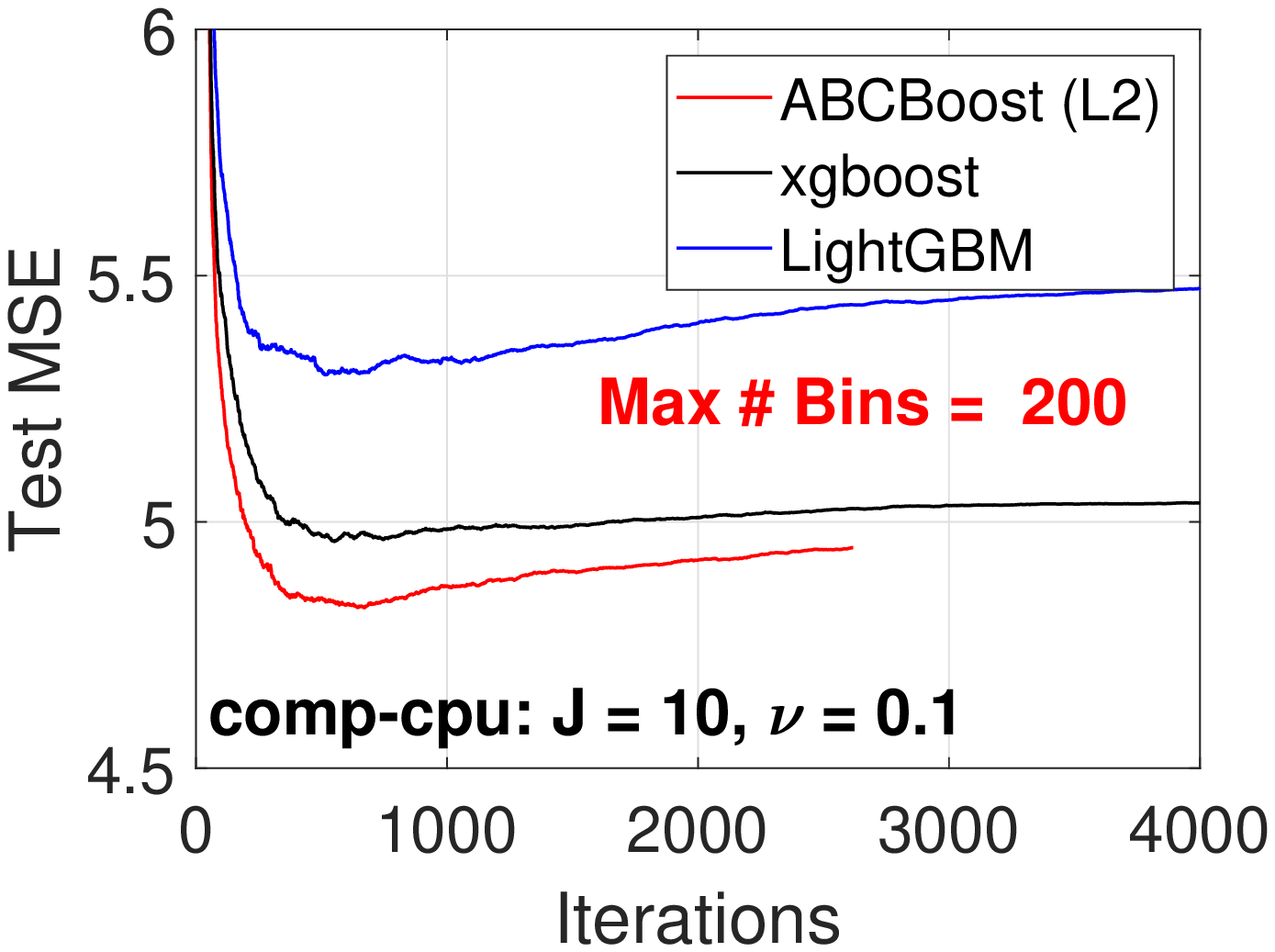}
}

\mbox{    
    \includegraphics[width=2.7in]{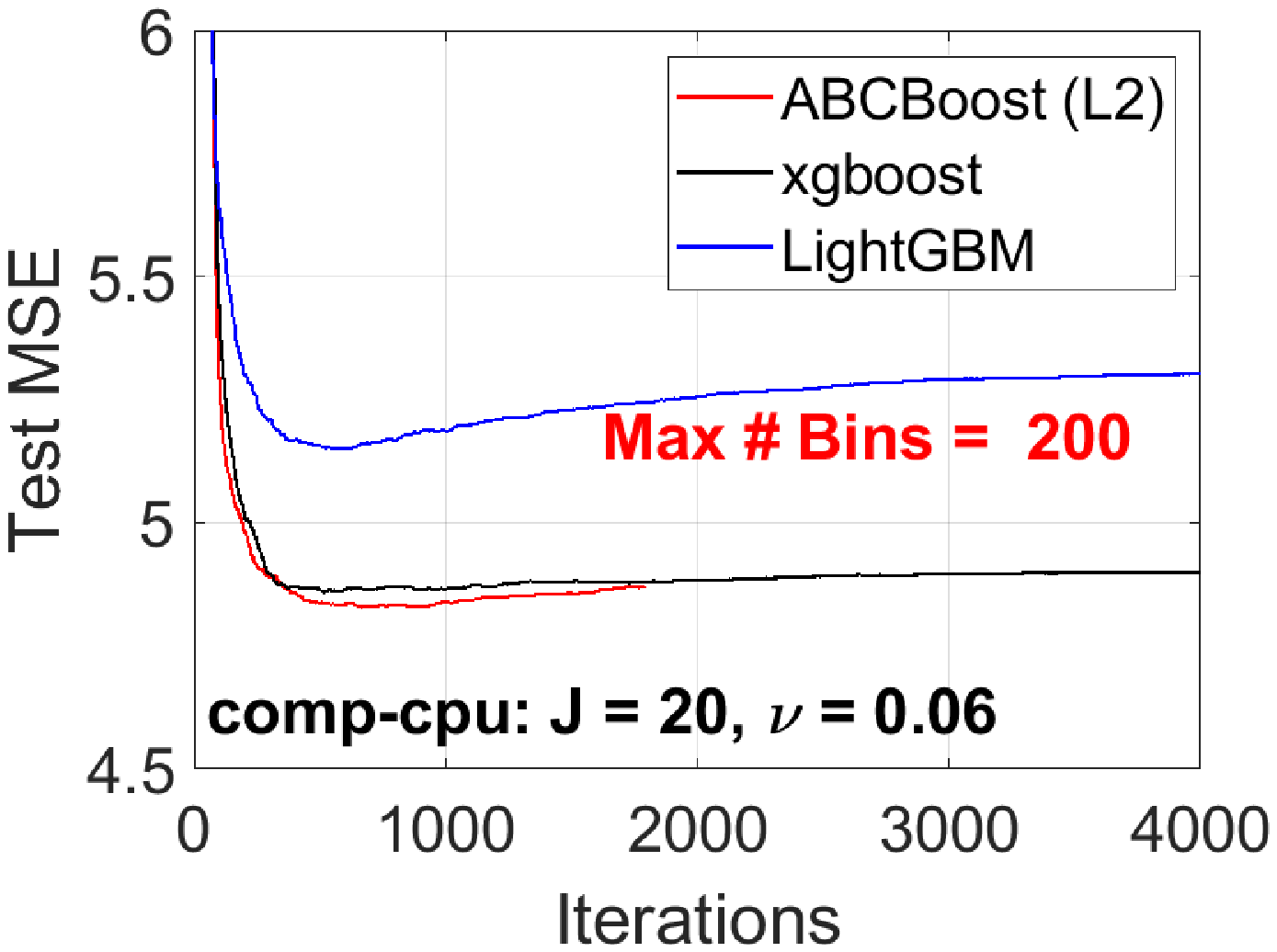}
    \includegraphics[width=2.7in]{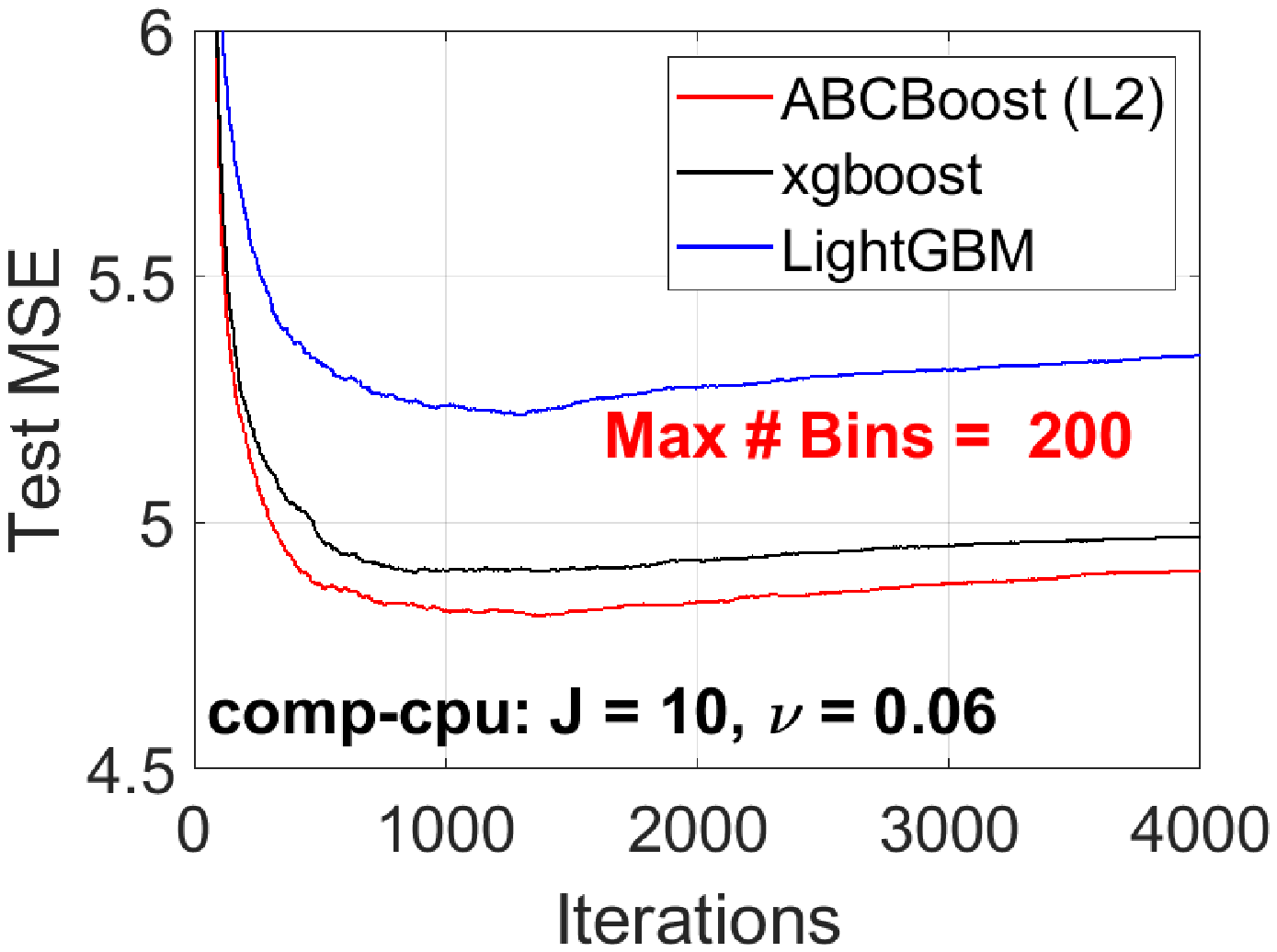}
}

\end{center}

\vspace{-0.15in}

    \caption{Test MSE history for all the iterations of $L_2$ boosting from three packages.}
    \label{fig:MSEHistoryThree}\vspace{-0.15in}
\end{figure}

Finally, Figure~\ref{fig:MSEHistoryThree} plots the test $L_2$ MSEs for all the iterations, at a particular set of parameters $J$, $\nu$, and MaxBin. Note that ABC-Boost package sets a conservative stopping criterion in Eq.~\eqref{eqn:stop}.

\newpage\clearpage

\section{Classification Using Robust LogitBoost}

Again, we denote a training dataset by $\{y_i,\mathbf{x}_i\}_{i=1}^n$, where $n$ is the number of training samples, $\mathbf{x}_i$ is the $i$-th feature vector, and  $y_i \in \{0, 1, 2, ..., K-1\}$ is the $i$-th class label, where $K=2$ for binary classification and $K\geq 3$ in multi-class classification. The class probabilities $p_{i,k}$  are assumed to be 
\begin{align}\label{eqn_logit}
p_{i,k} = \mathbf{Pr}\left(y_i = k|\mathbf{x}_i\right) = \frac{e^{F_{i,k}(\mathbf{x_i})}}{\sum_{s=0}^{K-1} e^{F_{i,s}(\mathbf{x_i})}},\hspace{0.2in} i = 1, 2, ..., n,
\end{align}
where $F$ is a function of $M$ terms:
\begin{align}\label{eqn_F_M}
F^{(M)}(\mathbf{x}) = \sum_{m=1}^M f_m
\end{align}
where  the base learner  $f_m$ is a regression tree, trained by minimizing the  negative log-likelihood loss:
\begin{align}\label{eqn_loss}
L = \sum_{i=1}^n L_i, \hspace{0.4in} L_i = - \sum_{k=0}^{K-1}r_{i,k}  \log p_{i,k}
\end{align}
where $r_{i,k} = 1$ if $y_i = k$ and $r_{i,k} = 0$ otherwise.  The optimization procedure requires  the first two derivatives of the loss function (\ref{eqn_loss}) with respective to the function values $F_{i,k}$ as follows: 
\begin{align}\label{eqn:logit_d1d2}
&\frac{\partial L_i}{\partial F_{i,k}} = - \left(r_{i,k} - p_{i,k}\right),
\hspace{0.5in}
\frac{\partial^2 L_i}{\partial F_{i,k}^2} = p_{i,k}\left(1-p_{i,k}\right),
\end{align}
which are standard results in textbooks.

\subsection{Tree-Splitting Criterion Using Second-Order Information}\label{sec_split}

Again, consider  a node with $N$ weights $w_i$, and $N$ response values $z_i$, $i=1$ to $N$, which are assumed to be ordered according to the sorted order of the corresponding feature values. The tree-splitting procedure seek the $t$ to maximize
\begin{align}\notag
Gain(t) =& \frac{\left[\sum_{i=1}^t z_iw_i\right]^2}{\sum_{i=1}^t w_i}+\frac{\left[\sum_{i=t+1}^N z_iw_i\right]^2}{\sum_{i=t+1}^{N} w_i}- \frac{\left[\sum_{i=1}^N z_iw_i\right]^2}{\sum_{i=1}^N w_i}
\end{align}
Plugging in  $w_i = p_{i,k}(1-p_{i,k})$, $z_i = \frac{r_{i,k}-p_{i,k}}{p_{i,k}(1-p_{i,k})}$  yields,
\begin{align}\label{eqn:logit_gain}
RLogitGain(t) =&  \frac{\left[\sum_{i=1}^t \left(r_{i,k} - p_{i,k}\right) \right]^2}{\sum_{i=1}^t p_{i,k}(1-p_{i,k})}+\frac{\left[\sum_{i=t+1}^N \left(r_{i,k}- p_{i,k}\right) \right]^2}{\sum_{i=t+1}^{N} p_{i,k}(1-p_{i,k})}- \frac{\left[\sum_{i=1}^N \left(r_{i,k} - p_{i,k}\right) \right]^2}{\sum_{i=1}^N p_{i,k}(1-p_{i,k})}.
\end{align}
Because the computations involve $\sum p_{i,k}(1-p_{i,k})$ as a group, this procedure is numerically robust/stable. This resolves the concerns in~\cite{friedman2000additive,friedman2001greedy,friedman2008response}.  In comparison, MART~\citep{friedman2001greedy}  used the first derivatives to construct  trees, i.e.,
\begin{align}\label{eqn:mart_gain}
MartGain(t) =&  \frac{1}{t}\left[\sum_{i=1}^t \left(r_{i,k} - p_{i,k}\right) \right]^2+
\frac{1}{N-t}\left[\sum_{i=t+1}^N \left(r_{i,k} - p_{i,k}\right) \right]^2-\frac{1}{N}
\left[\sum_{i=1}^N \left(r_{i,k} - p_{i,k}\right) \right]^2.
\end{align}

{\scriptsize\begin{algorithm}{\small
$F_{i,k} = 0$, $p_{i,k} = \frac{1}{K}$, $k = 0$ to  $K-1$, $i = 1$ to $n$ \\
For $m=1$ to $M$ Do\\
\hspace{0.1in}    For $k=0$ to $K-1$ Do\\
\hspace{0.2in}  $\left\{R_{j,k,m}\right\}_{j=1}^J = J$-terminal node regression tree from
 $\{r_{i,k} - p_{i,k}, \ \ \mathbf{x}_{i}\}_{i=1}^n$,  with weights $p_{i,k}(1-p_{i,k})$, using the tree split gain formula Eq.~\eqref{eqn:logit_gain}.\\
 \hspace{0.2in}  $\beta_{j,k,m} = \frac{K-1}{K}\frac{ \sum_{\mathbf{x}_i \in
  R_{j,k,m}} r_{i,k} - p_{i,k}}{ \sum_{\mathbf{x}_i\in
  R_{j,k,m}}\left(1-p_{i,k}\right)p_{i,k} }$ \\
\hspace{0.2in}  $F_{i,k} = F_{i,k} +
\nu\sum_{j=1}^J\beta_{j,k,m}1_{\mathbf{x}_i\in R_{j,k,m}}$ \\
 \hspace{0.1in} End\\
\hspace{0.12in} $p_{i,k} = \exp(F_{i,k})/\sum_{s=0}^{K-1}\exp(F_{i,s})$\\
End
\caption{Robust LogitBoost. MART is similar, with the only difference in Line 4.  }
\label{alg:robust_LogitBoost}}
\end{algorithm}}

Algorithm~\ref{alg:robust_LogitBoost} describes Robust LogitBoost using the tree split gain formula in Eq.~\eqref{eqn:logit_gain}. Note that after trees are constructed, the values of the terminal nodes are computed by
\begin{align}\notag
\frac{\sum_{node} z_{i,k} w_{i,k}}{\sum_{node} w_{i,k}} =
\frac{\sum_{node} \left(r_{i,k} - p_{i,k}\right)}{\sum_{node} p_{i,k}(1-p_{i,k})},
\end{align}
which explains Line 5 of Algorithm~\ref{alg:robust_LogitBoost}. For MART~\citep{friedman2001greedy}, the algorithm is almost identical to Algorithm~\ref{alg:robust_LogitBoost}, except for Line 4, which for MART uses the tree split gain formula Eq.~\eqref{eqn:mart_gain}.

Note that, for binary classification (i.e., $K=2$), we just need to build one tree per iteration. 

\subsection{Experiments on Binary Classification}

The binary classification dataset, ``ijcnn1'', with 49990 training samples  and 91701 testing samples, is available at
\url{https://www.csie.ntu.edu.tw/~cjlin/libsvmtools/datasets/binary.html} . This dataset was used in a competition and LIBSVM (kernel SVM) was the winner which achieved a test error of 1293 (out of 91701 test examples).

Again, from the terminal, the following command 
{\scriptsize
\begin{verbatim}
./abcboost_train -method robustlogit -data data/ijcnn1.train.csv -J 20 -v 0.1 -iter 10000  -data_max_n_bins 1000
\end{verbatim}
}
\noindent trains a binary classification model with ``Robust LogitBoost'' algorithm using $J=20$ leaf nodes, $\nu = 0.1$ shrinkage, and $M=10000$ iterations, with MaxBin = 1000. Two files are created:
\begin{verbatim}
ijcnn1.train.csv_robustlogit_J20_v0.1.model
ijcnn1.train.csv_robustlogit_J20_v0.1.trainlog
\end{verbatim}
\noindent We print out the first 3 rows and last 3 rows of the ``.trainlog'' file:
\begin{verbatim}
   1 3.06638501851082e+04    2475 0.00956
   2 2.73916764152665e+04    2342 0.00897
   3 2.46653571044355e+04    2336 0.01209
  
9998 8.65973959207622e-14       0 0.00829
9999 9.05941988094128e-14       0 0.00842
10000 8.92619311798626e-14       0 0.00844  
\end{verbatim}
\noindent where the second column is the training loss, the third  is the training error, and the third column is the wall-clock time. Again, to ensure to output deterministic results, we train all cases with single-thread.  Here, we would like to emphasize that multi-threaded programs may be possible to output even better (but non-deterministic) results, due to the potentially beneficial stochastic effect. 

\newpage 

The next command 
{\scriptsize
\begin{verbatim}
./abcboost_predict -data data/ijcnn1.test.csv -model ijcnn1.train.csv_robustlogit_J20_v0.1.model 
\end{verbatim}
}
\noindent outputs the test results 
\begin{verbatim}
ijcnn1.test.csv_robustlogit_J20_v0.1.prediction
ijcnn1.test.csv_robustlogit_J20_v0.1.testlog    
\end{verbatim}
\noindent The last 10 rows of the ``.testlog'' file are shown below: 
\begin{verbatim}
9991 1.70802827618969e+04    1179 0.00306
9992 1.70918813774666e+04    1182 0.00312
9993 1.70853256019377e+04    1181 0.00321
9994 1.71042704318109e+04    1182 0.00299
9995 1.70905462461684e+04    1180 0.00291
9996 1.71042853137969e+04    1182 0.00306
9997 1.70978375521247e+04    1181 0.00284
9998 1.70948331080451e+04    1180 0.00243
9999 1.70970460294472e+04    1182 0.00317
10000 1.71022010098107e+04    1180 0.00270    
\end{verbatim}
where the third column records the test errors. Recall the best result of LIBSVM was 1293. Note that in the Appendix of~\citet{li2010robust}, the experimental results on ijcnn1 dataset were also presented. 

\vspace{0.2in}

\begin{figure}[h]

\vspace{0.2in}
\begin{center}
\mbox{    
    \includegraphics[width=2.9in]{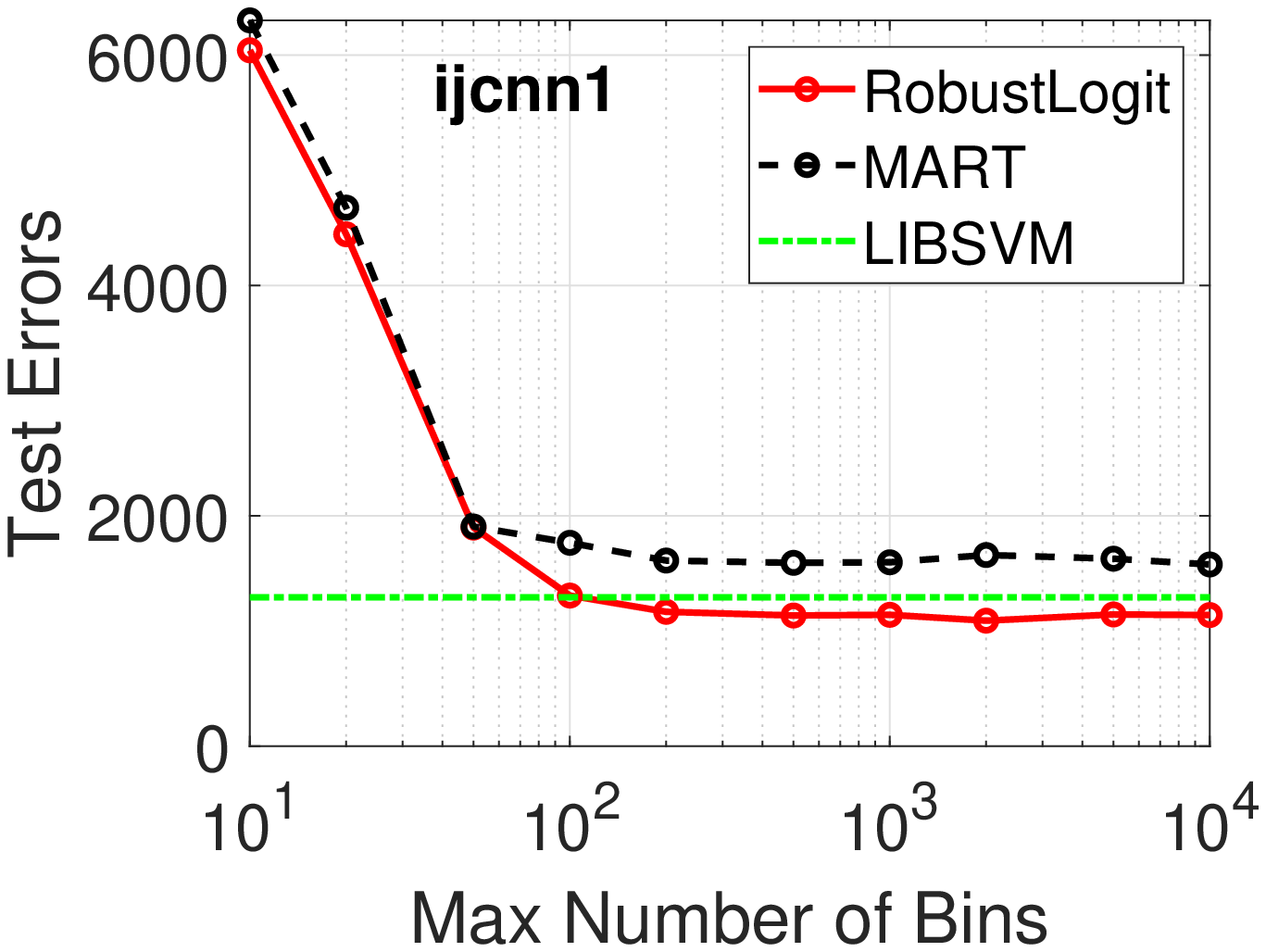}
    \includegraphics[width=2.9in]{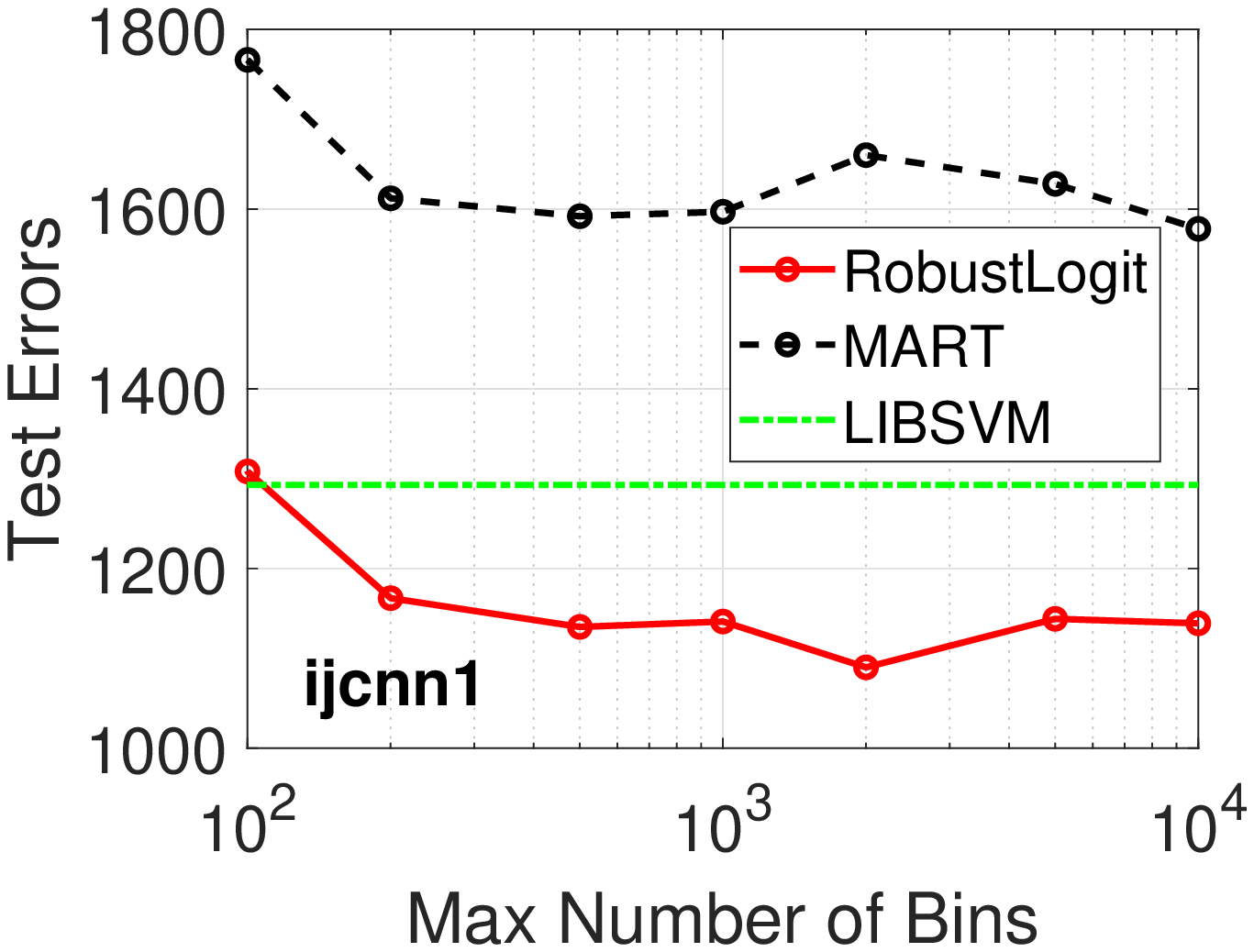}    
}

\end{center}

\vspace{-0.1in}

    \caption{Best Test errors on the ijcnn1 dataset, to compare Robust LogitBoost with MART, with respect to MaxBin (maximum number of bins), which ranges from 10 to $10^4$. This dataset was used in a competition and LIBSVM was the winner (with the best test error being 1293).  The right panel is merely the zoomed-in version of the left panel.}
    \label{fig:Error_ijcnn1}
\end{figure}

Figure~\ref{fig:Error_ijcnn1} plots the best test errors to compare Robust LogitBoost with MART, for $J\in\{10, 20\}$, $\nu \in \{0.06, 0.1\}$, and $M=10000$. The results are presented with respect to MaxBin (max number of bins) to illustrate the impact of binning on the classification errors. Indeed, the simple fixed-length binning algorithm we implemented in this package does not perform well when MaxBin is set to smaller than 100. Again, this is expected. 

\newpage

\begin{figure}[h]
\begin{center}

\mbox{    
    \includegraphics[width=2.9in]{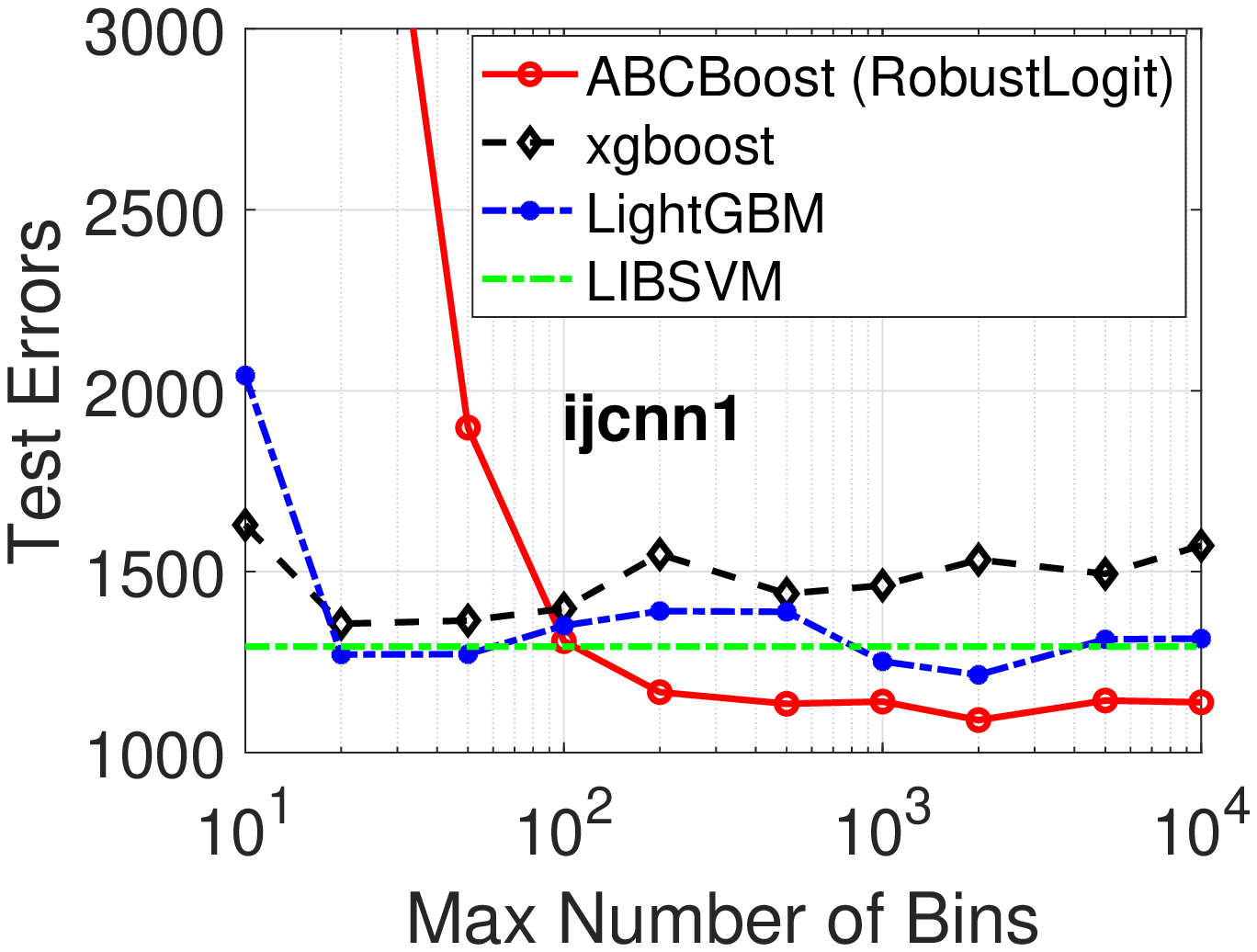}
    \includegraphics[width=2.9in]{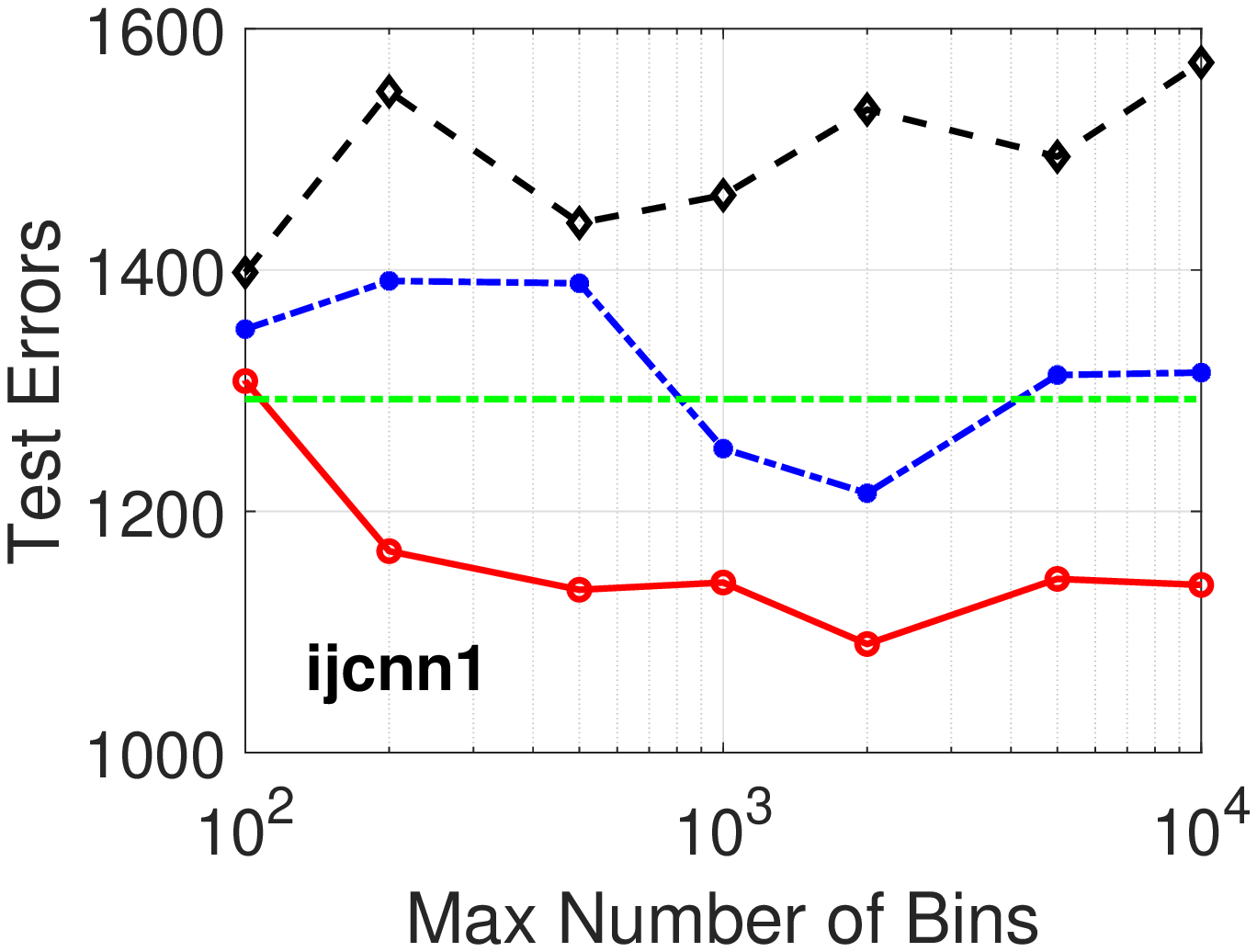}    
}

\end{center}

\vspace{-0.2in}

    \caption{Best test errors of all three boosting packages, as well as the best LIBSVM result. We can see that the simple fixed-length binning method used in our package does not perform so well when MaxBin is set to be smaller than 100. }
    \label{fig:error_ijcnn1_three}\vspace{0.2in}
\end{figure}

Figure~\ref{fig:error_ijcnn1_three} compares Robust LogitBoost with xgboost and LightGBM at the same MaxBin values. Clearly, when MaxBin is set to be smaller than 100,  the simple fixed-length binning method implemented  in our package does not perform well. On the other hand, the best (lowest) errors are attained at MaxBin chosen to be much larger than 100 (in fact 2000 for this dataset). 

While we do expect there is considerable room to improve the simple (fixed-length) binning algorithm, we admit that we haven't found one that is universally (or largely) better, after using this binning scheme for about 15 years. 

\vspace{0.1in}

\begin{figure}[h]
\begin{center}
\mbox{    
    \includegraphics[width=2.7in]{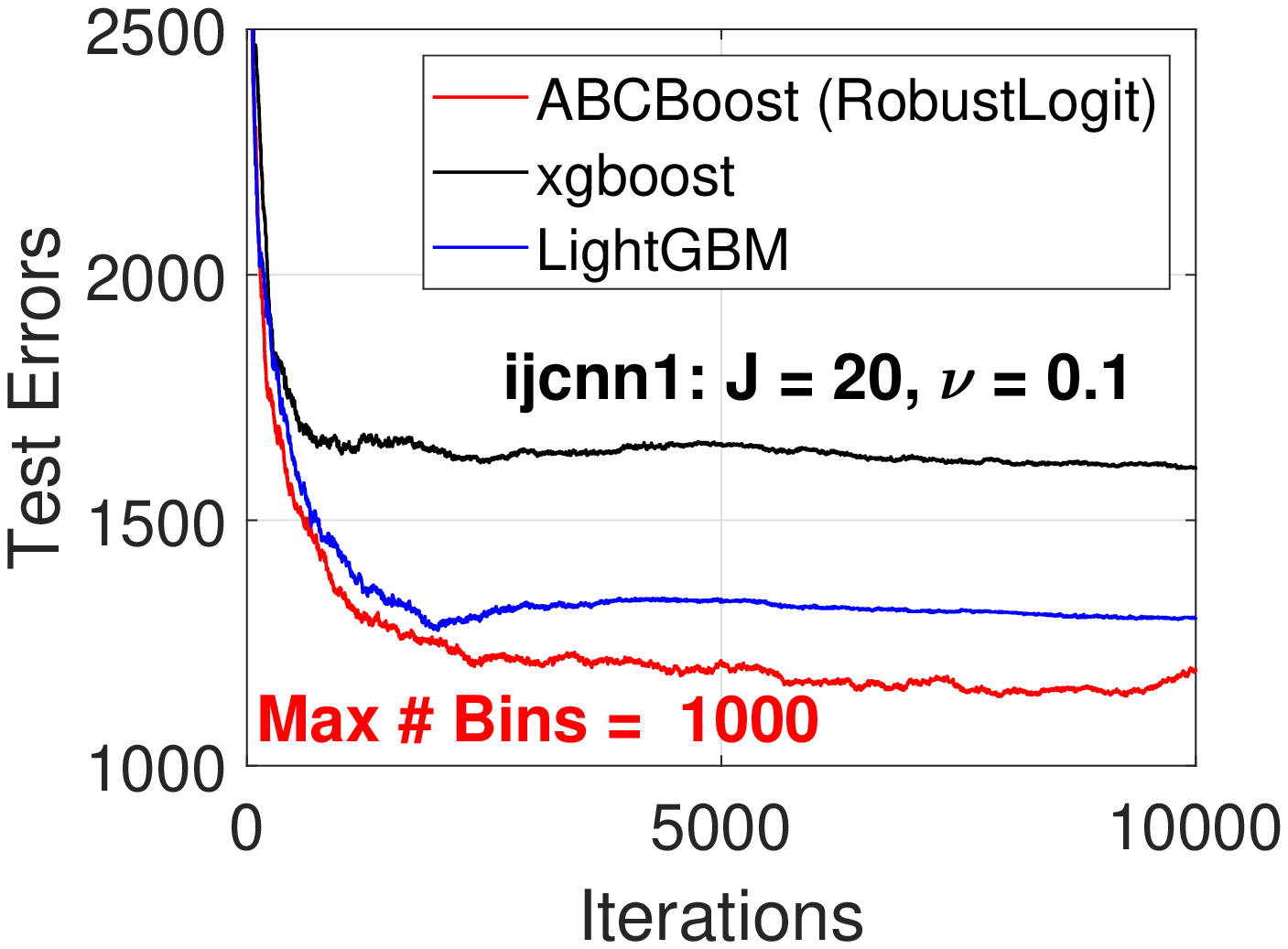}
    \includegraphics[width=2.7in]{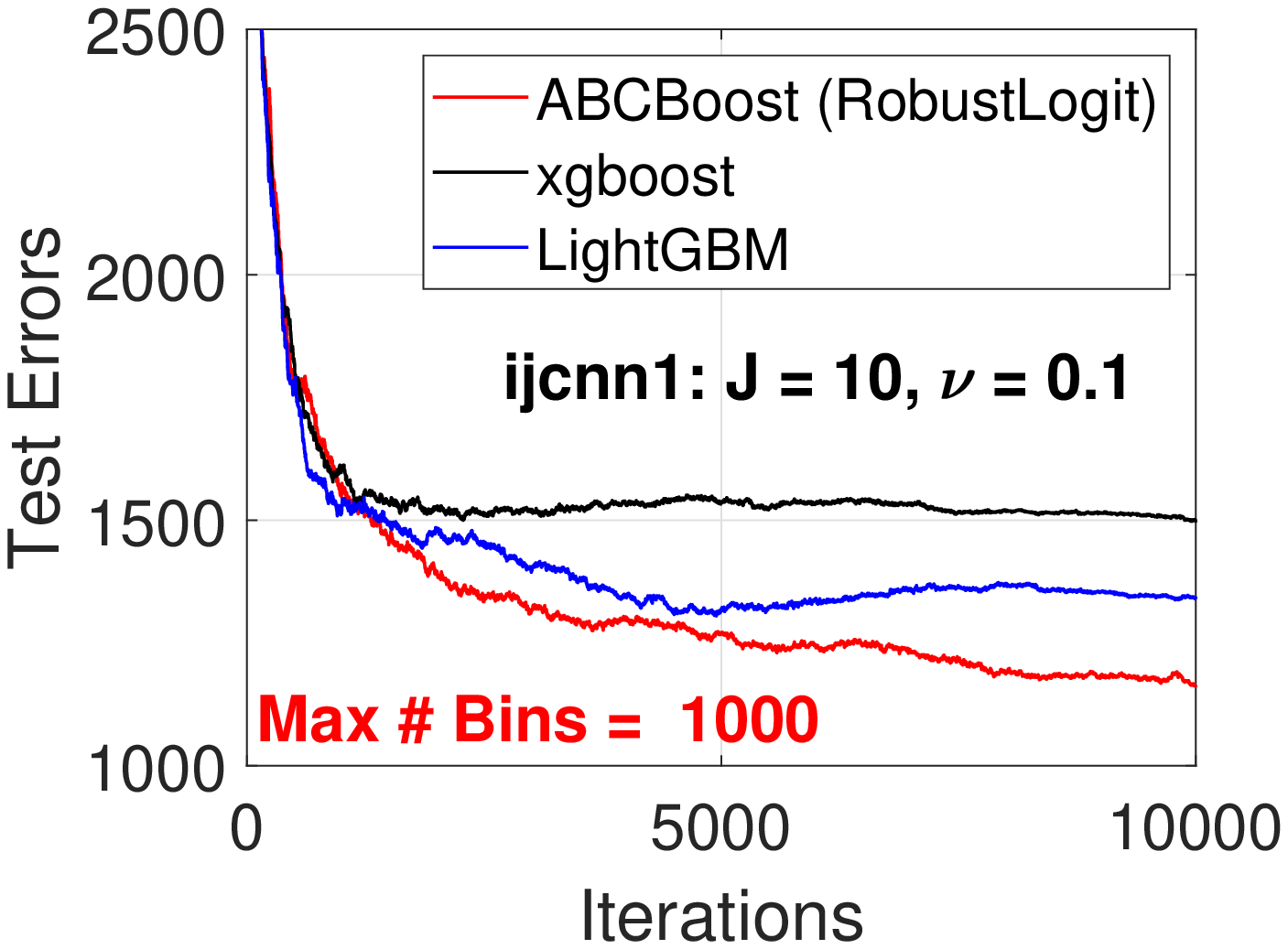}
}

\mbox{    
    \includegraphics[width=2.7in]{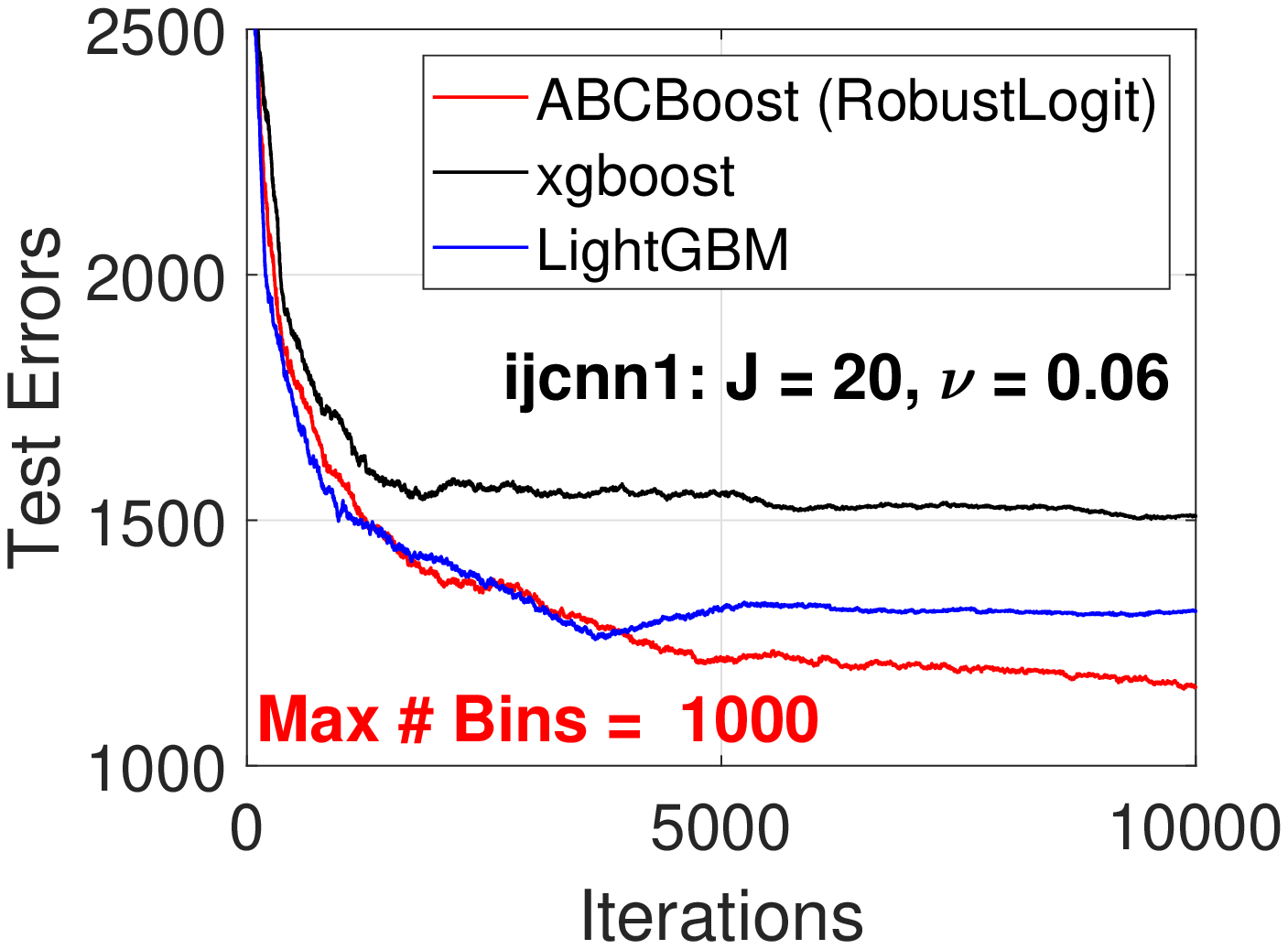}
    \includegraphics[width=2.7in]{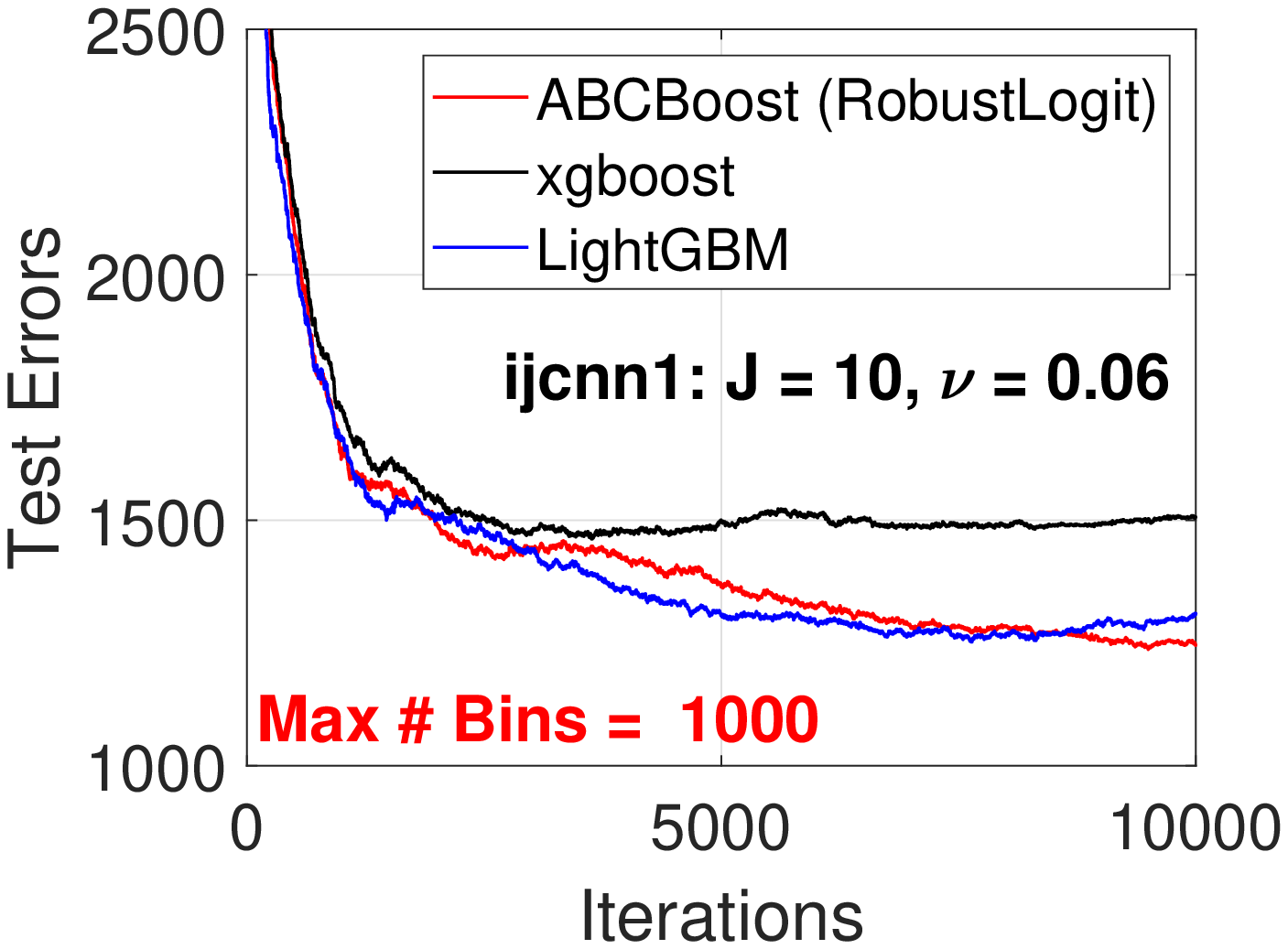}
}

\mbox{    
    \includegraphics[width=2.7in]{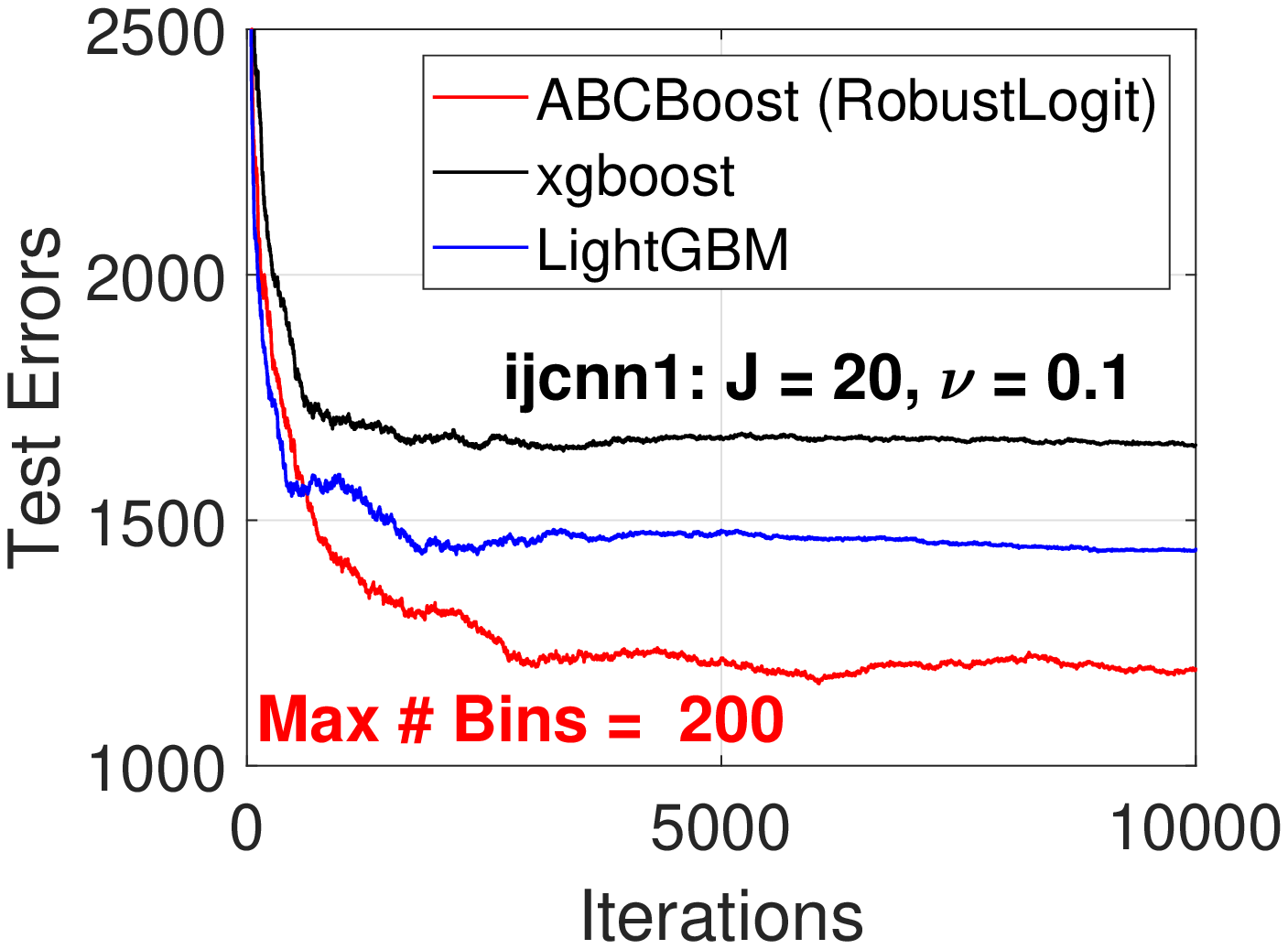}
    \includegraphics[width=2.7in]{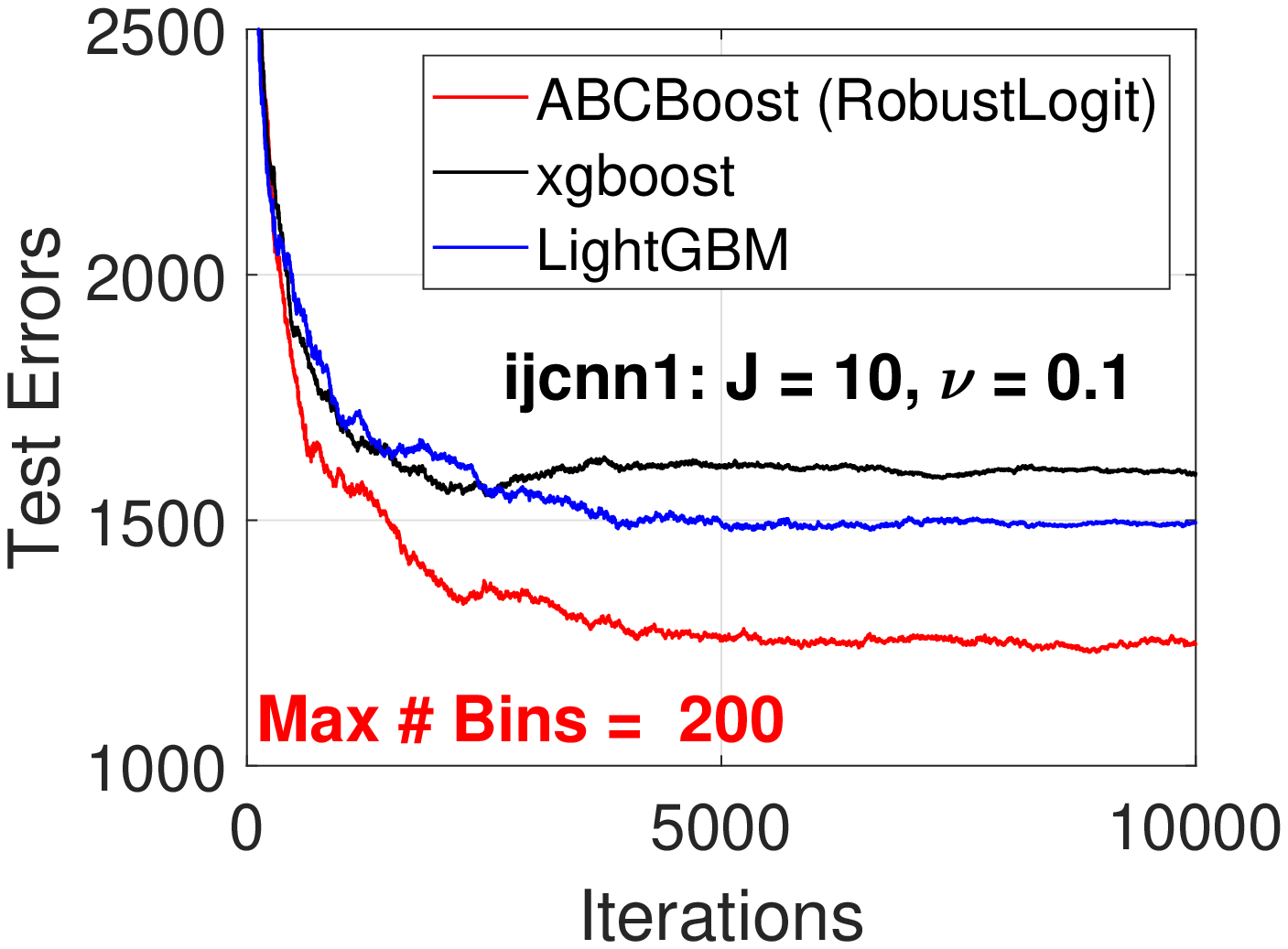}
}

\mbox{    
    \includegraphics[width=2.7in]{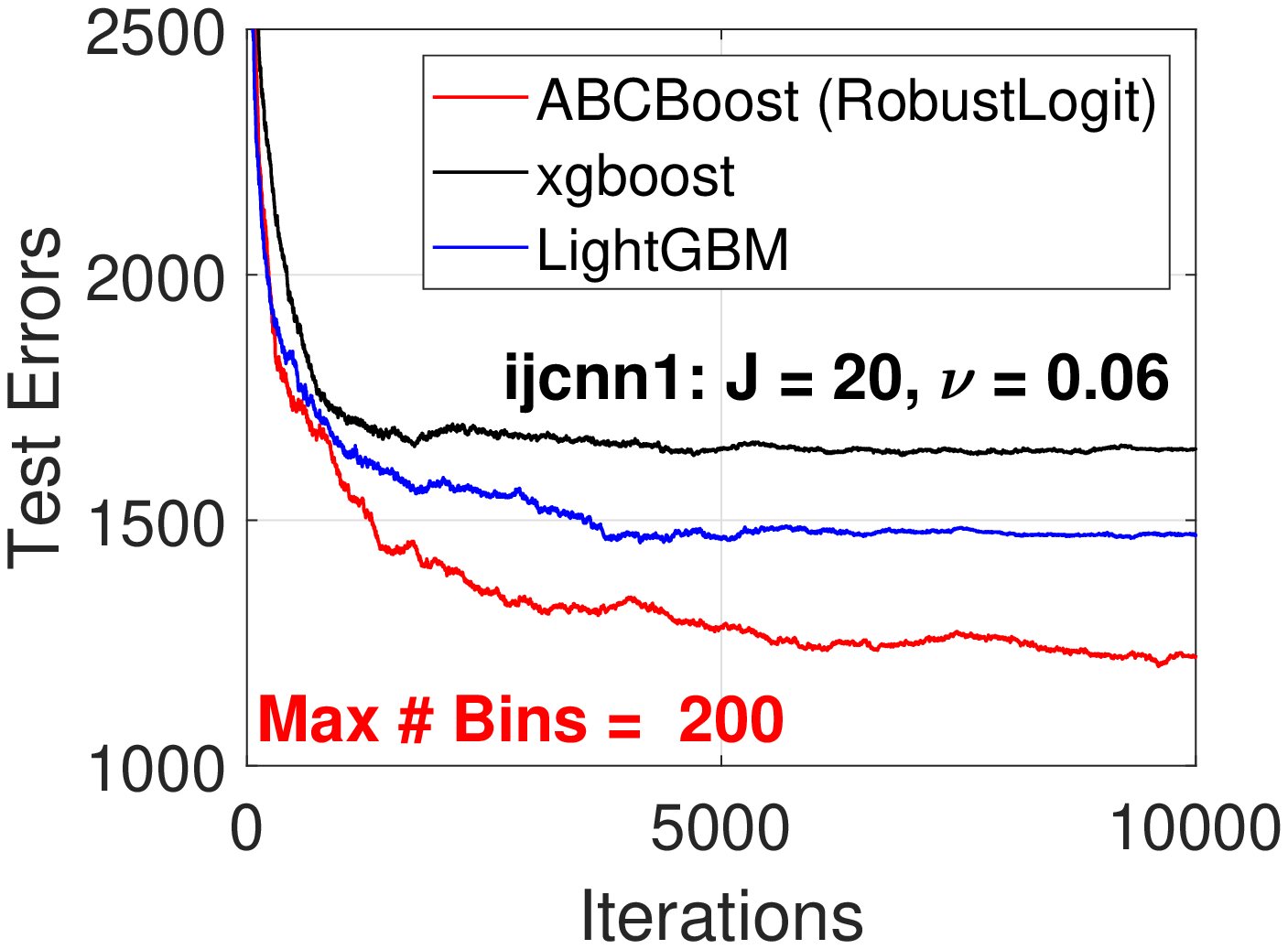}
    \includegraphics[width=2.7in]{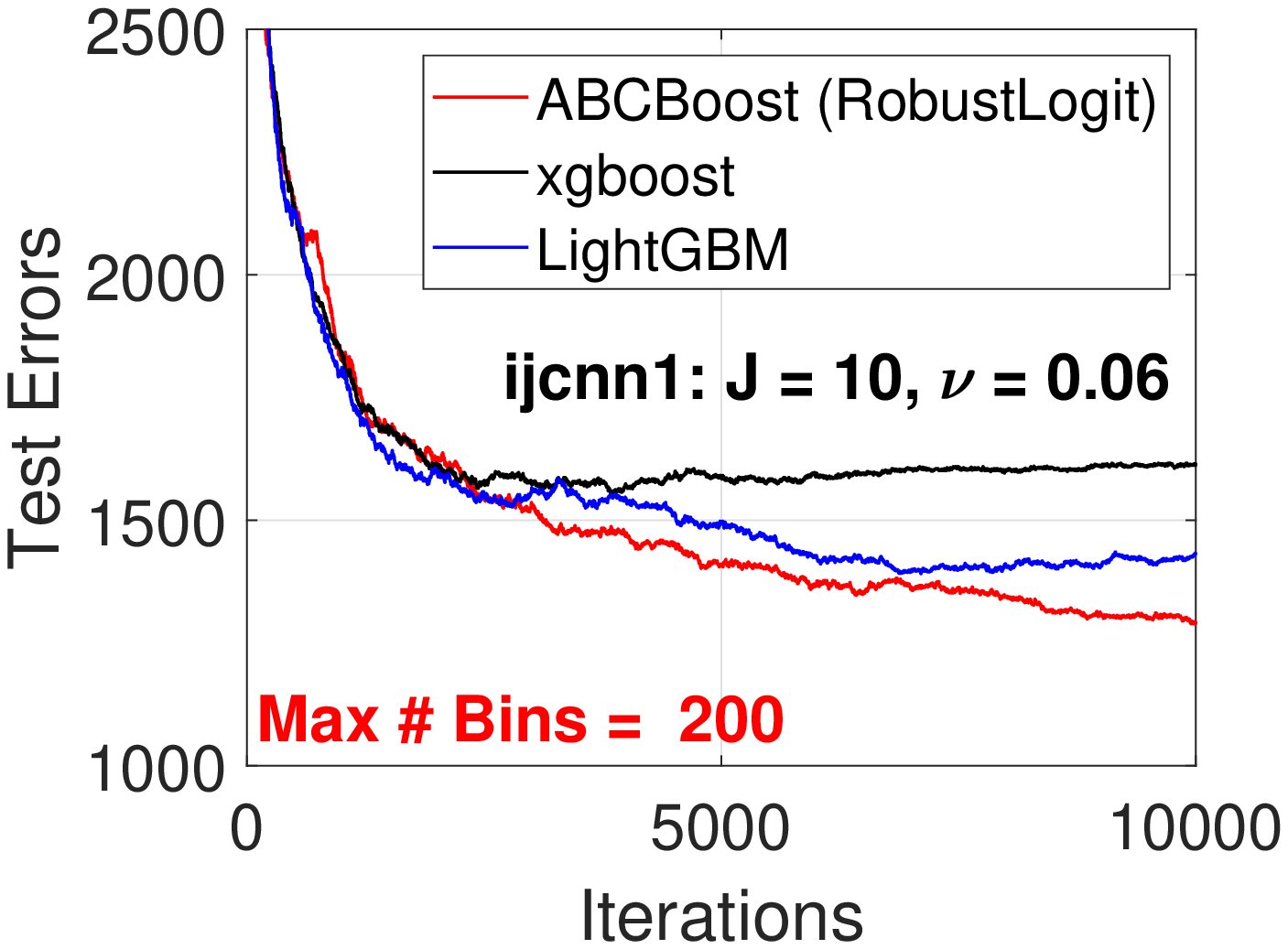}
}

\end{center}

    \caption{Test error history for all the iterations, at particular set of parameters ($J$, $\nu$, MaxBin), to compare RobustLogitBoost as implemented in our package with xgboost and LightGBM.}
    \label{fig:errorHistoryThree_ijcnn1}
\end{figure}

Finally, in Figure~\ref{fig:errorHistoryThree_ijcnn1} we plot the history of test errors for all $M=10000$ iterations for each set of parameters ($J$, $\nu$, MaxBin), to compare RobustLogitBoost with xgboost and LightGBM. We hope that it is clear from the plots that practitioners might want to re-visit this simple binning method, to further better understand why it works so well and to further improve its performance. 

\newpage\clearpage

\section{ABC-Boost for Multi-Class Classification}

The idea of ``adaptive base class boost'' (ABC-Boost) was originated from~\citep{li2008adaptive}, in which the classical (textbook) derivatives of the multi-class logistic regression were re-written to be
\begin{align}\label{eqn:abc_d1}
&\frac{\partial L_i}{\partial F_{i,k}}  = \left(r_{i,0} - p_{i,0}\right) - \left(r_{i,k} - p_{i,k}\right),\\\label{eqn:abc_d2}
&\frac{\partial^2 L_i}{\partial F_{i,k}^2} = p_{i,0}(1-p_{i,0}) + p_{i,k}(1-p_{i,k}) + 2p_{i,0}p_{i,k},
\end{align}
where, we recall that, 
\begin{align}\label{eqn_logit}
&L = \sum_{i=1}^N L_i, \hspace{0.4in} L_i = - \sum_{k=0}^{K-1}r_{i,k}  \log p_{i,k}\\
&p_{i,k} = \mathbf{Pr}\left(y_i = k|\mathbf{x}_i\right) = \frac{e^{F_{i,k}(\mathbf{x_i})}}{\sum_{s=0}^{K-1} e^{F_{i,s}(\mathbf{x_i})}}
\end{align}
In the above we have assume class 0 is the ``base class'' and used the ``sum-to-zero'' constraint on the $F_i$ values. In the actual implementation, we will need to identify the base class at each iteration. 

\vspace{0.1in}

As shown  in~\cite{li2009abc,li2010robust}, the ``exhaustive search'' strategy works well in term of accuracy but it is highly inefficient. The unpublished technical report by~\citet{li2008adaptive} proposed the ``worst-class'' search strategy and the other unpublished report by~\citet{li2010fast} proposed the ``gap'' strategy. Very recently, ~\citet{li2022fast} developed a unified framework to achieve ``Fast ABC-Boost''   by introducing three parameters: (i) The ``search'' parameters $s$ restricts the search for the base class within the $s$-worst classes. (ii) The ``gap'' parameter $g$ indicates that the search for base class is only conducted at every $g+1$ iterations. (iii) Finally the ``warm-up'' parameter $w$ specifies that the search only starts after we have trained Robust LogitBoost or MART for $w$ iterations. 

\vspace{0.1in}

Algorithm~\ref{alg:fast-abc-LogitBoost} summarizes the unified framework of Fast ABC-Boost. Although it introduces additional parameters $(s,g,w)$, the good  news is that in most cases the performance is not sensitive to these parameters. In fact, the ``worst-class'' strategy as initially developed in~\citet{li2008adaptive} already works pretty well, although for some cases it might result in ``catastrophic failures''. In a sense, these parameters $(s,g,w)$ are introduced mainly to avoid ``catastrophic failures''. 

In Algorithm~\ref{alg:fast-abc-LogitBoost}, the gain formula for tree split in ABC-RobustLogitBoost is similar to~\eqref{eqn:logit_gain}:
\begin{align}\label{eqn:abclogit_gain}
ABCRLogitGain(t,b) =&  \frac{\left[\sum_{i=1}^t \left(r_{i,k} - p_{i,k}\right) -\left(r_{i,b} - p_{i,b}\right)\right]^2}{\sum_{i=1}^t p_{i,b}(1-p_{i,b})+p_{i,k}(1-p_{i,k})+2p_{i,b}p_{i,k}}\\\notag
&+\frac{\left[\sum_{i=t+1}^N \left(r_{i,k} - p_{i,k}\right) -\left(r_{i,b} - p_{i,b}\right)\right]^2}{\sum_{i=t+1}^N p_{i,b}(1-p_{i,b})+p_{i,k}(1-p_{i,k})+2p_{i,b}p_{i,k}}\\\notag
&-\frac{\left[\sum_{i=1}^N \left(r_{i,k} - p_{i,k}\right) -\left(r_{i,b} - p_{i,b}\right)\right]^2}{\sum_{i=1}^N p_{i,b}(1-p_{i,b})+p_{i,k}(1-p_{i,k})+2p_{i,b}p_{i,k}}
\end{align}

\begin{algorithm}[h]{\small
$F_{i,k} = 0$,\ \  $p_{i,k} = \frac{1}{K}$, \ \ \ $k = 0$ to  $K-1$, \ $i = 1$ to $n$ \\
$L_{\textit{prev}}^{(k)} = \sum_{i=1}^{N}{1_{y_i = k}}$, \ \ \ $k = 0$ to $K - 1$\\
For $m=1$ to $M$ Do\\
\hspace{0.1in} If $(m - 1) \textit{ mod } (g + 1) = 0$ Then\\
\hspace{0.2in}    $L_{\textit{prev}}^{(k'_0)}, L_{\textit{prev}}^{(k'_1)}, \dots, L_{\textit{prev}}^{(k'_{K-1})} = \text{Sort} \  L_{\textit{prev}}^{(k)}, \ k = 0$ to $K-1$ decreasingly.\\
\hspace{0.2in}    $\textit{search\_classes} = \{k'_0, k'_1, \dots, k'_{s-1}\}$\\
\hspace{0.1in} Else\\
\hspace{0.2in} $\textit{search\_classes} = \{B(m - 1)\}$\\
\hspace{0.1in} End\\
\hspace{0.1in}    For $b \in \textit{search\_classes}$, Do\\
\hspace{0.2in}    For $k=0$ to $K-1$, $k\neq b$, Do\\
\hspace{0.3in}  $\left\{R_{j,k,m}\right\}_{j=1}^J = J$-terminal
node regression tree from  $\{-(r_{i,b} - p_{i,b}) +  (r_{i,k} - p_{i,k}), \ \ \mathbf{x}_{i}\}_{i=1}^n$  with weights $p_{i,b}(1-p_{i,b})+p_{i,k}(1-p_{i,k})+2p_{i,b}p_{i,k}$, using the tree split gain formula Eq.~\eqref{eqn:abclogit_gain}.
 \\
 \hspace{0.3in}  $\beta_{j,k,m} = \frac{ \sum_{\mathbf{x}_i \in
  R_{j,k,m}} -(r_{i,b} - p_{i,b}) + (r_{i,k} - p_{i,k})  }{ \sum_{\mathbf{x}_i\in
  R_{j,k,m}} p_{i,b}(1-p_{i,b})+ p_{i,k}\left(1-p_{i,k}\right) + 2p_{i,b}p_{i,k} }$ \\
\hspace{0.3in}  $G_{i,k,b} = F_{i,k} +
\nu\sum_{j=1}^J\beta_{j,k,m}1_{\mathbf{x}_i\in R_{j,k,m}}$ \\
 \hspace{0.2in} End\\
\hspace{0.2in} $G_{i,b,b} = - \sum_{k\neq b} G_{i,k,b}$ \\
\hspace{0.2in}  $q_{i,k} = \exp(G_{i,k,b})/\sum_{s=0}^{K-1}\exp(G_{i,s,b})$ \\
\hspace{0.2in} $L^{(b)} = -\sum_{i=1}^N \sum_{k=0}^{K-1} r_{i,k}\log\left(q_{i,k}\right)$\\
\hspace{0.1in} End\\
\hspace{0.1in} $B(m) = \underset{b}{\text{argmin}} \  \ L^{(b)}$\\
\hspace{0.1in} $F_{i,k} = G_{i,k,B(m)}$\\
\hspace{0.1in}  $p_{i,k} = \exp(F_{i,k})/\sum_{s=0}^{K-1}\exp(F_{i,s})$ \\
End}
\caption{Fast-ABC-RobustLogitBoost using the ``$s$-worst classes'' search strategy and the ``gap‘’ strategy (with parameter $g$) for the base class. }
\label{alg:fast-abc-LogitBoost}
\end{algorithm}

\newpage\clearpage

\begin{figure}[h]
\begin{center}

\mbox{    
    \includegraphics[width=4in]{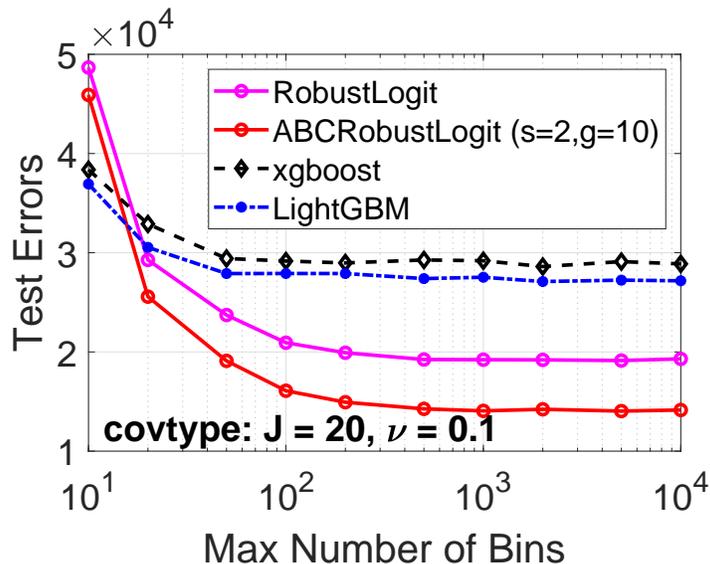}
}

\end{center}

\vspace{-0.2in}

    \caption{Test errors for $J=20$ and $\nu=0.1$ on the covtype dataset, for comparing four methods. Note that xgboost and LightGBM are supposed to be the same as Robust LogitBoost. We suspect the performance differences are caused by different binning schemes. }
    \label{fig:covtypeErrorJ20v01Four}
\end{figure}

We use the UCI ``covtype'' dataset for the demonstration. This dataset contains 581012 examples and we split half/half for training/testing. It is a 7-class classification problem. In the experiment, we let $J=20$, $\nu=0.1$, and $M=1000$. The following commands: 

{\scriptsize
\begin{verbatim}
./abcboost_train -method abcrobustlogit -data data/covtype.train.csv -J 20 -v 0.1 -iter 1000 -search 2 -gap 10
./abcboost_predict -data data/covtype.test.csv -model covtype.train.csv_abcrobustlogit2g10_J20_v0.1_w0.model     
\end{verbatim}
}
\noindent train and test  ``ABC RobustLogitBoost'' with $s=2$ and $g=10$.  We can of course also train the regular Robust LogitBoost as in Algorithm~\ref{alg:robust_LogitBoost}. Figure~\ref{fig:covtypeErrorJ20v01Four} compares the test errors for four different methods: Robust LogitBoost, ABC Robust LogitBoost, xgboost, and LightGBM, for binning parameters MaxBin ranging from 10 to $10^4$.

Figure~\ref{fig:covtypeErrorJ20v01Four} shows that the simple fixed-length binning scheme implemented in our package does not perform well when MaxBin = 10. This is expected. For this multi-class classification task, it is clear that ``ABC Robust LogitBoost'' improves ``Robust LogitBoost'' quite considerably.

\vspace{0.1in}

\begin{figure}[t]
\begin{center}
\mbox{    
    \includegraphics[width=2.7in]{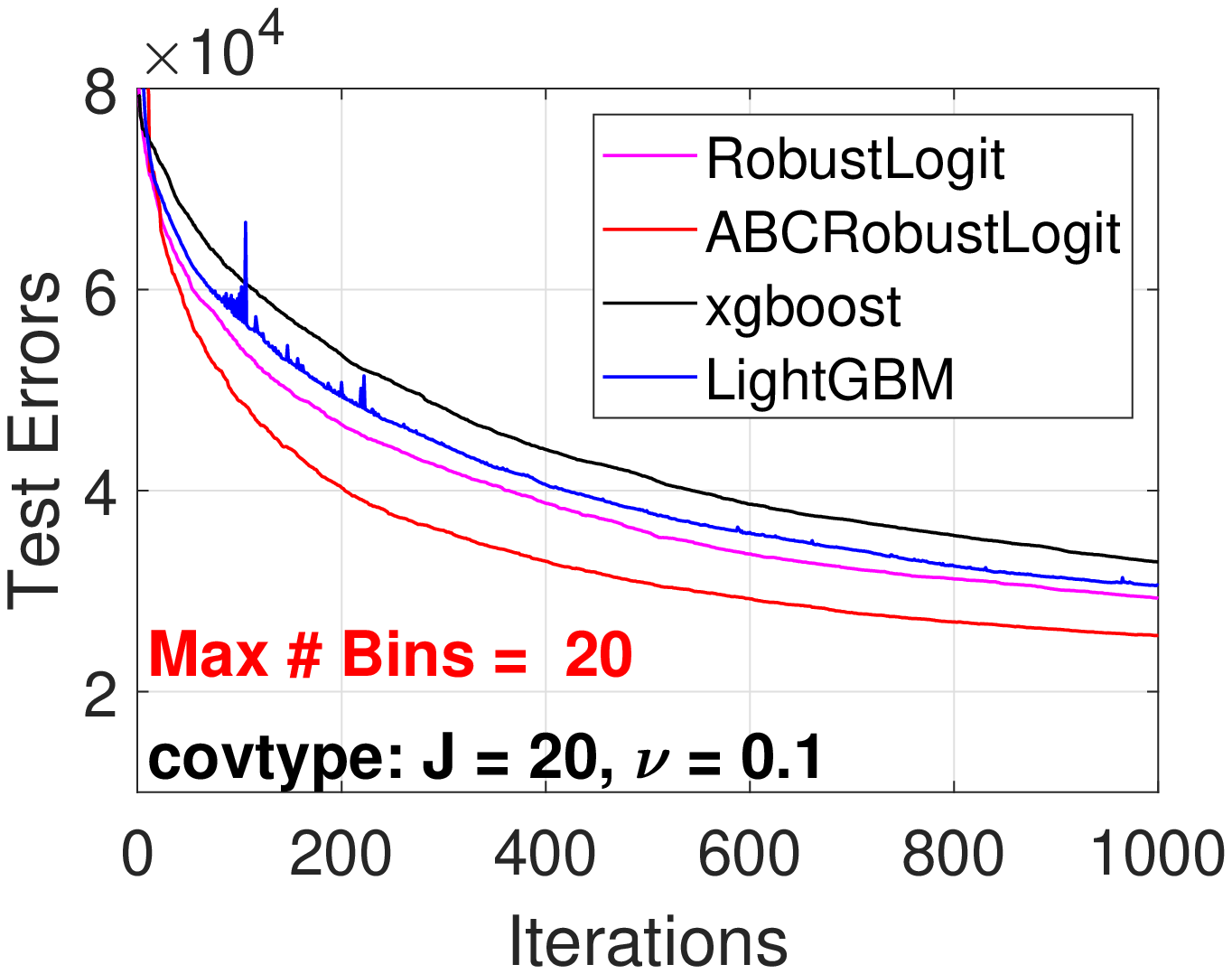}
    \includegraphics[width=2.7in]{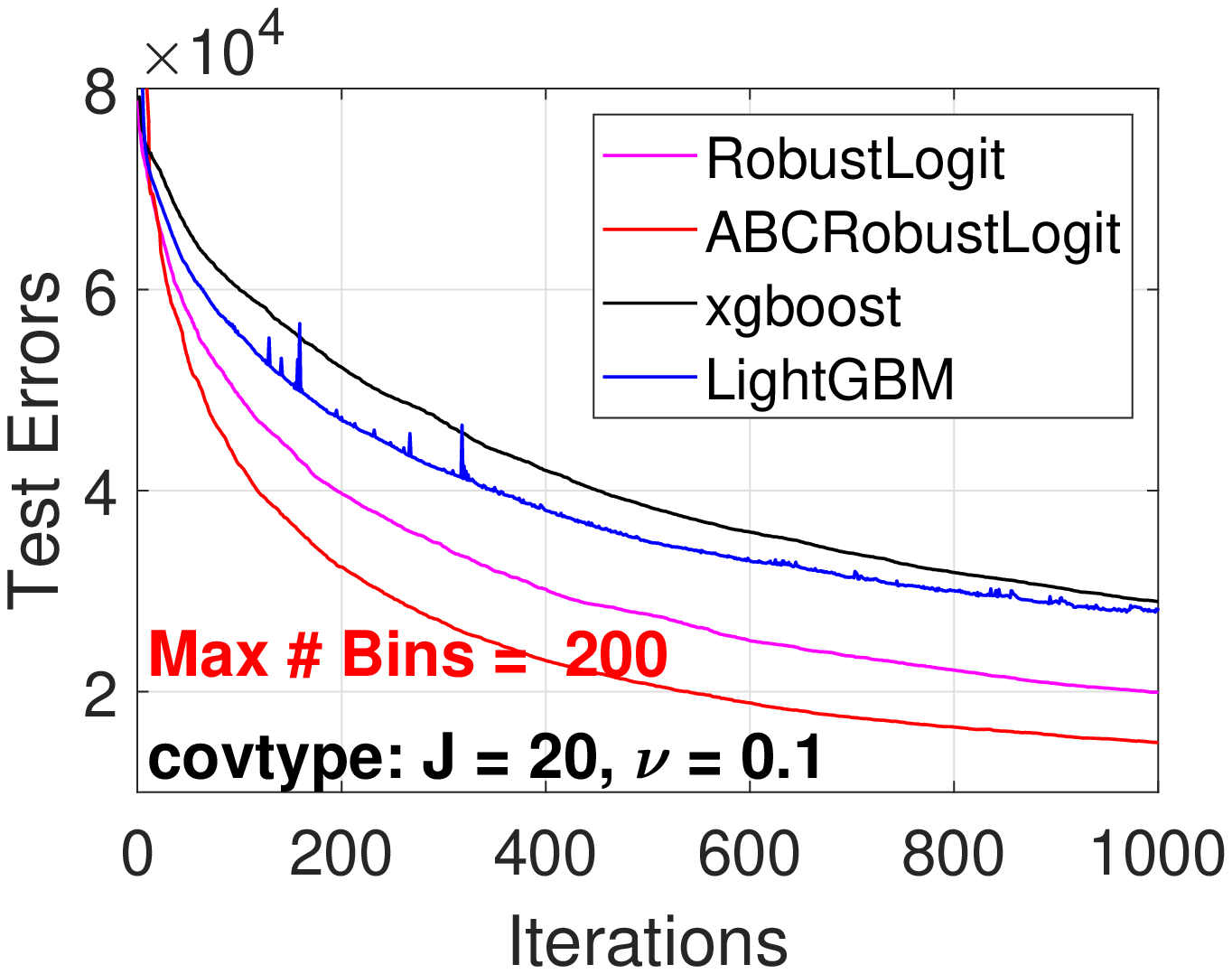}
}

\mbox{    
    \includegraphics[width=2.7in]{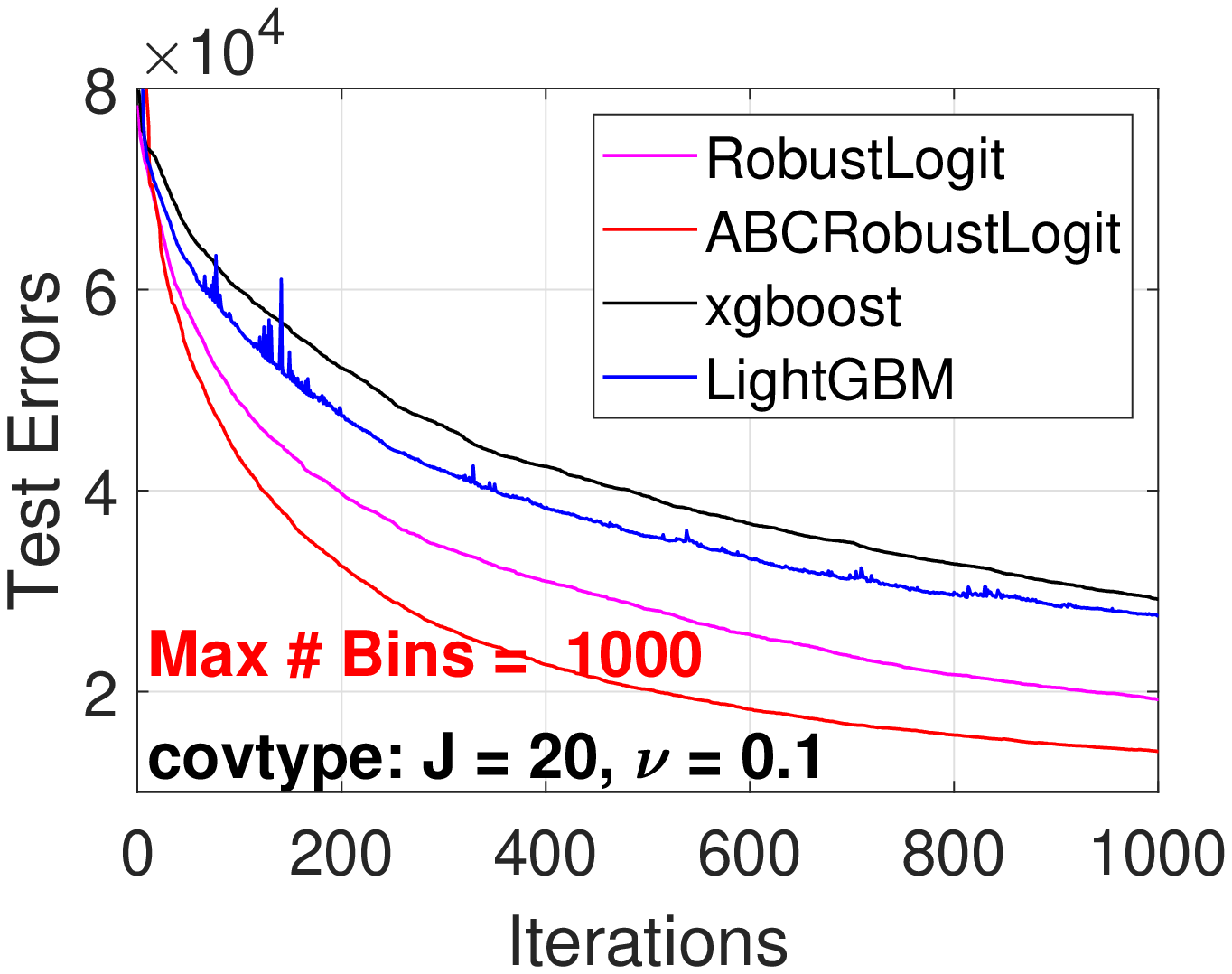}
    \includegraphics[width=2.7in]{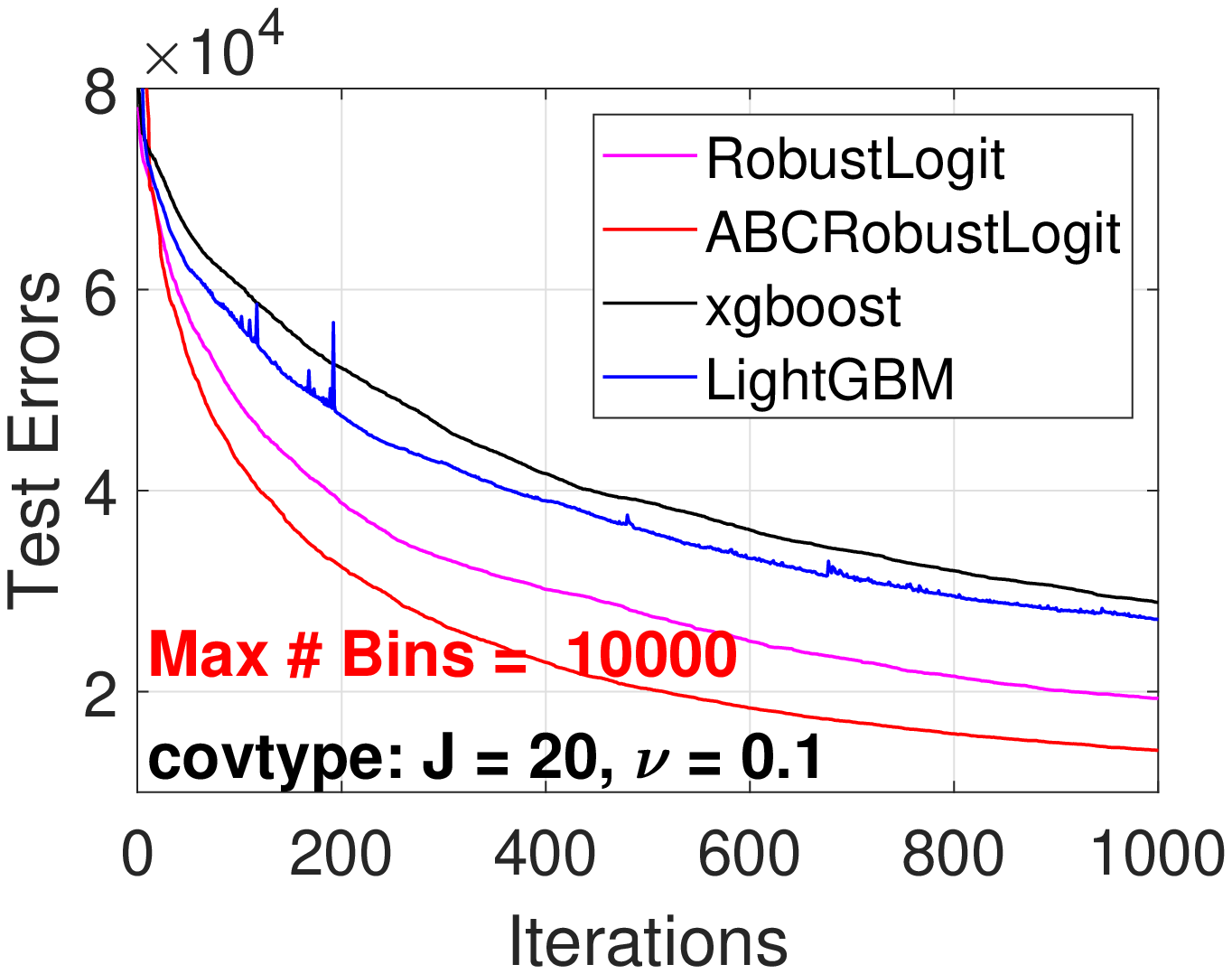}
}

\end{center}

\vspace{-0.25in}

    \caption{Test error histories for the UCI covtype dataset, to compare four methods at different MaxBin values, for $J=20$ and $\nu=0.1$.  }
    \label{fig:errorHistoryFour_covtype}\vspace{-0.05in}
\end{figure}

Finally, Figure~\ref{fig:errorHistoryFour_covtype} plots the histories of test errors for four methods. 

\newpage

\vspace{-0.05in}
\section{Conclusion}
\vspace{-0.05in}

A decade ago (or  earlier), the author(s) finished the contributions on boosting and trees in~\citet{li2007mcrank,li2008adaptive,li2009abc,li2010fast,li2010robust}, which generated a lot of interests (see for example  discussions  \url{https://hunch.net/?p=1467}  in 2010) and motivated the developments of popular boosted tree platforms using (i) feature-binning; (ii) second-order gain formula for tree splitting, as the standard implementation.  The authors then shifted interests to other  topics including deep neural networks and computational advertising for commercial search engines; see e.g.,~\citet{fan2019mobius,zhao2020distributed,fei2021gemnn,xu2021agile,zhao2022communication}. While we use deep neural networks extensively for commercial ads applications, we witness that boosted trees are still very popular in industry. Our own ``rule-of-thumb'' is that one should first try boosted trees if the applications have less than 10000 (handcrafted or pre-generated) features and less than 100 million training examples. 

\vspace{0.1in}

\noindent We notice that ``adaptive base class boost'' (ABC-Boost) has not become part of the popular boosted tree platforms. This is probably because in the formally published papers~\citep{li2009abc,li2010robust}, we only reported the computationally expensive exhaustive search strategy, which might prevent practitioners from trying ABC-Boost. We thus decide to release  ``Fast ABC-Boost'' which is really the result of the efforts in the past 15 years as summarized in~\citet{li2022fast}.

\vspace{0.1in}

\noindent Finally, we should mention that in the implementation, the  ``best-first'' tree-growing strategy, is always used, in all our papers on boosting and trees including~\citet{li2007mcrank}. Ping Li got this idea from attending Professor Jerry Friedman's class (and being his TA) in early 2000's.  See pages 76 -- 77 of the tutorial \url{http://www.stat.rutgers.edu/home/pingli/doc/PingLiTutorial.pdf} , which was compiled and edited while the authors worked at Cornell University and Rutgers University.

\newpage\clearpage

\bibliographystyle{plainnat}
\bibliography{scholar_refs}

\end{document}